\documentclass[10pt,journal,compsoc]{IEEEtran}
\usepackage{amsmath,amsfonts}
\usepackage{amsthm}

\usepackage{amsthm}
\usepackage{algorithm}
\usepackage{algorithmic}
\usepackage{array}
\usepackage[caption=false,font=normalsize,labelfont=sf,textfont=sf]{subfig}
\usepackage{textcomp}
\usepackage{stfloats}
\usepackage{url}
\usepackage{verbatim}
\usepackage{graphicx}

\usepackage{graphicx}
\usepackage{amsmath}
\usepackage{amssymb}
\usepackage{booktabs}
\usepackage{colortbl, xcolor}
\usepackage{booktabs,caption,fixltx2e}
\usepackage[flushleft]{threeparttable}
\usepackage{multirow} 
\usepackage{array}
\usepackage{booktabs}
\usepackage{multirow}
\usepackage{xcolor}
\usepackage{xcolor}
\definecolor{dropred}{RGB}{163,45,45}
\newcommand{\cellval}[2]{#1\%\;{\textcolor{dropred}{\small$\downarrow$\scriptsize#2}}}
\usepackage{makecell}
\usepackage{xcolor}
\definecolor{cvprgold}{RGB}{214, 120, 0}      
\definecolor{cvprblue}{RGB}{0, 130, 60}      
\usepackage{amssymb}   
\usepackage{pifont}    
\newcommand{\cmark}{\checkmark} 
\newcommand{\xmark}{\ding{55}}  
\newcommand{\etal}{\textit{et al.}}

\newcommand{\methodname}{LLM-Demorph}
\ifCLASSOPTIONcompsoc
  \usepackage[nocompress]{cite}
\else
  \usepackage{cite}
\fi
\ifCLASSINFOpdf
\else
\fi
\begin{document}
%
\title{Enhancing Single-Image Facial Demorphing using Multimodal Large Language Models}
%
%
%
%

\author{Nitish~Shukla,~\IEEEmembership{Member,~IEEE,}
        Arun~Ross,~\IEEEmembership{Senior Member,~IEEE}
\thanks{Manuscript received April 19, 2026; revised --- --, ----.}}

%
%

\markboth{Journal of \LaTeX\ Class Files,~Vol.~14, No.~8, August~2015}%
{Shell \MakeLowercase{\textit{et al.}}: Bare Demo of IEEEtran.cls for Computer Society Journals}
%



\IEEEtitleabstractindextext{%
\begin{abstract}
Face recognition systems are increasingly vulnerable to morphing attacks, where a composite image is crafted to match multiple identities, enabling unauthorized access and identity fraud. Existing detection methods identify morphed images but cannot recover constituent images or identities, limiting their forensic utility. This paper presents a novel reference-free facial demorphing framework that leverages Multimodal Large Language Models (MLLMs) to guide a coupled diffusion-based reconstruction process. Our key innovation lies in extracting semantic embeddings from intermediate MLLM layers to condition the demorphing, providing high-level reasoning about facial attributes and identity cues that complement low-level pixel information. We formulate demorphing as a coupled conditional generation problem, where both constituent faces are synthesized jointly through a denoising diffusion model operating directly in the RGB domain, ensuring inter-identity consistency while preserving fine-grained perceptual details. Unlike prior approaches that rely on compressed latent representations or assume identity overlap between training and testing sets, our method bypasses lossy text generation-reencoding cycles by directly utilizing MLLM hidden states as conditioning signals, enabling the denoising network to attend to subtle visual cues such as hair, background, and facial textures. Extensive experiments across multiple benchmarks spanning landmark-based and learning-based morphing techniques demonstrate state-of-the-art performance: restoration accuracies exceeding 96\% at 0.1\% False Match Rate (FMR) on landmark-based morphs, and superior performance on challenging StyleGAN morphs, with PSNR gains exceeding 6--9~dB over existing methods. Ablation studies further reveal that middle MLLM layers encode more identity-discriminative representations, RGB-domain demorphing outperforms latent-space approaches by 30--40\% at strict operating points, and full MLLM embeddings provide substantial advantages over raw ViT features through enhanced semantic structuring from multimodal pretraining.
\end{abstract}

}

\maketitle

\IEEEdisplaynontitleabstractindextext

%
\IEEEpeerreviewmaketitle

\section{Introduction}

\IEEEPARstart{A}{utomated}  face recognition has become a useful technology in modern security infrastructure \cite{boutros2022elasticface,deng2019arcface,11395540}, serving as the primary authentication mechanism in applications ranging from border control to smartphone unlocking. However, the integrity of face recognition systems is  threatened by a number of attacks, including face morphing attacks, wherein a composite image is crafted to simultaneously match multiple identities, enabling identity fraud and passport forgery \cite{DBLP:conf/icb/FerraraFM14,caldeira2023unveiling,damer2023mordiff,DBLP:journals/tbbis/ZhangVRRDB21,DBLP:conf/visapp/MakrushinND17}.

The proliferation of sophisticated morphing techniques has considerably escalated this threat. While early landmark-based methods \cite{DBLP:conf/icb/FerraraFM14,DBLP:conf/visapp/MakrushinND17,quek2019facemorpher} relied on geometric warping and pixel-level blending, recent approachehukla leverage generative models such as StyleGAN \cite{stylegan}, diffusion models \cite{10744444}, and flow-based networks \cite{bizzi2025flowing} to produce visually convincing morphs that preserve identity characteristics from both contributors, making them increasingly difficult for  Morph Attack Detection (MAD) systems to identify.

Research has primarily addressed this threat through Single-image MAD (S-MAD) \cite{DBLP:conf/visapp/MakrushinND17,8486607,9190629,8014962,DBLP:conf/wacv/CaldeiraOCIPBSD25,neto2022orthomad,caldeira2023unveiling}, which determines whether a given image is morphed, and Differential MAD (D-MAD) \cite{damer2019detecting,10.1007/978-3-319-94211-7_48,9067912,9091520}, which exploits a trusted reference image along with the input image for detection. While these approaches have shown promising results, they fundamentally address only the \emph{recognition} of morphed images rather than the \emph{reconstruction} of constituent images or identities. Face demorphing inverts the morphing process to recover original faces, offering a more comprehensive solution that additionally enables forensic analysis and identity verification.
\begin{figure}
    \centering
    \includegraphics[width=\linewidth]{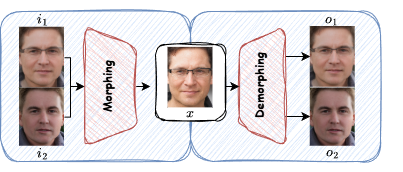}
    \caption{Overview of the morphing and demorphing process. A morphing operator blends two face images $i_1$ and $i_2$ to create the morph image $x$ (Left). A demorphing operator decomposes $x$ to recover the reconstructed images $o_1$ and $o_2$ from the morph (Right). }
    \label{fig:overview_main}
\end{figure}
Similar to MAD, demorphing can be categorized into \emph{differential (reference-based)} \cite{8730323,10415238} and \emph{single-image reference-free} \cite{diffdemorph,ipd,lcgan,dcgan} settings. In the differential setting, one constituent image (usually a trusted live capture  at a kiosk, for example) is available alongside the morph, significantly constraining the solution space. The reference-free (RF) setting is substantially more challenging: demorphing becomes a one-to-many ill-posed inverse problem requiring decomposition into two plausible faces over the non-linear manifold of facial variations encompassing identity, expression, age, and pose, etc. Existing approaches using adversarial learning \cite{dcgan,ref51}, latent space optimization \cite{lcgan}, and diffusion-based reconstruction \cite{diffdemorph} often struggle to simultaneously achieve high identity fidelity and perceptual quality, particularly on GAN or diffusion-generated morphs.

In this work, we reframe face demorphing as a semantically-guided conditional generation problem. Our key insight is that Multimodal Large Language Models (MLLMs), which have demonstrated remarkable zero-shot reasoning capabilities across diverse visual domains, can provide rich semantic guidance for the demorphing process. We extract hidden state representations from intermediate MLLM layers encoding high-level facial attributes, identity cues, and structural characteristics, then use these embeddings to condition a coupled diffusion model that jointly generates both constituent faces. By providing the morph in the RGB domain alongside semantic conditioning, the denoising network can attend to fine-grained perceptual details often lost in compressed representations, while the coupled formulation ensures both reconstructed identities are generated interdependently, preserving the inherent relationship established by the morphing process.

We validate our method on multiple benchmark datasets spanning landmark-based and learning-based morphing techniques. On landmark-based morphs, the method achieves restoration accuracies exceeding 96\% at 0.1\% False Match Rate (FMR). On generative morphs, our method outperforms the closest competitor by $+7.01$ percentage points at $0.1\%$ FMR on challenging StyleGAN morphs and by $+8.18$ points in Rank-1 retrieval. Comprehensive ablation studies further reveal the critical role of MLLM semantic guidance and the impact of different architectural choices on demorphing performance.

Our contributions are as follows:

\begin{itemize}
    \item We propose a \textbf{prompt-driven, reference-free facial demorphing framework} leveraging MLLMs to inject high-level semantic reasoning into the demorphing process.
    
    \item We formulate demorphing as a \textbf{coupled conditional diffusion problem}, enabling joint reconstruction of both constituent identities from a single morph.
    
    \item We introduce a \textbf{novel conditioning strategy} utilizing intermediate MLLM hidden states, combining semantic guidance with RGB-domain morph conditioning.
    
    \item We provide \textbf{extensive ablation analysis} across MLLM embedding depths (ViT, initial, middle, last) and conditioning strategies, yielding insights into how abstraction levels influence demorphing.
    
    \item Evaluations on multiple benchmarks demonstrate \textbf{state-of-the-art performance}, particularly at strict FMR operating points and on challenging GAN and diffusion-based morphs.
\end{itemize}
\begin{figure}
    \centering
    \includegraphics[width=\linewidth]{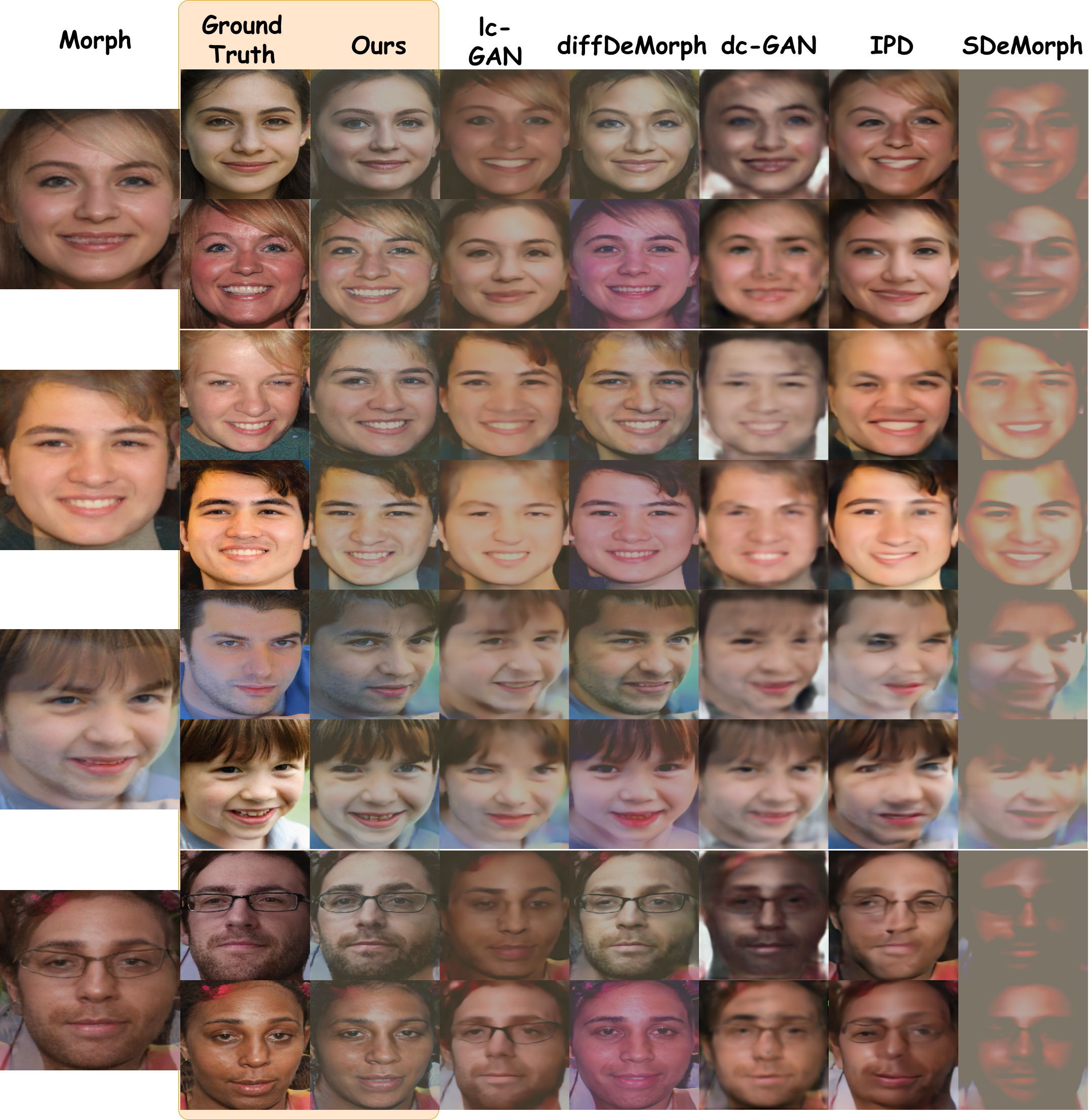}
    \caption{Reconstruction results from our LLM-guided demorphing approach.
Our method successfully generates visually distinct identity reconstructions while maintaining high fidelity to the corresponding ground-truth images.}
    \label{fig:main-result}
\end{figure}

\section{Preliminaries}
\subsection{Face Morphing}
A face morph is generated by blending identity features from multiple facial images of distinct individuals to produce a single image that inherits biometric characteristics from each contributing identity. Such morphs pose a significant threat to face recognition systems, as they can simultaneously match multiple 
identities, enabling unauthorized access or identity fraud \cite{DBLP:conf/icb/FerraraFM14}. Contemporary face morphing methods can be categorized into two groups; landmark-based generation and learning-based generation.
\subsubsection{Landmark-Based morph Generation} 
Landmark-based face morphing methods first detect key facial points (e.g., eyes, nose, mouth) from the source images and compute the corresponding points of the morphed face via linear interpolation. Delaunay triangulation \cite{Lee1980} is then applied to perform affine warping of the original images, which are subsequently blended at the pixel level to generate the final morph. 
Early work by Ferrara et al. \cite{DBLP:conf/icb/FerraraFM14} employed GIMP/GAP \cite{GIMP,GIMP2} for manual morph creation. Markrushin et al. \cite{DBLP:conf/visapp/MakrushinND17} introduced automated complete and splicing morphs, enabling large-scale morph generation. Modern tools, such as FaceMorph \cite{quek2019facemorpher}, use STASM \cite{7294955}  for landmark detection and Delaunay triangulation \cite{Lee1980} for warping and blending, but rely on pixel averaging outside the facial region, producing artifacts around hair and neck areas. OpenCV-based methods \cite{openCVmorph,SMDD} combine dlib \cite{10.5555/1577069.1755843} landmarks with additional triangulation. Despite the addition of triangles outside the facial region, artifacts are still present in the generated morphed image.

\subsubsection{Learning-based Morph Generation}
Learning-based morph generation methods produce morphed facial images by interpolating the latent representations of two source faces. Damer et al. \cite{damer2018morgan} proposed an encoder–decoder–discriminator network that generates morphs via adversarial learning. Venkatesh et al. \cite{DBLP:conf/iwbf/VenkateshZRRDB20} leveraged the pre-trained StyleGAN \cite{stylegan} generator to synthesize morphed faces, achieving high visual quality but often compromising the identity and structural integrity of the contributors. Building on this, Zhang et al. \cite{DBLP:journals/tbbis/ZhangVRRDB21} introduced an identity and structure-preserving optimization loss to better retain the original facial characteristics. More recently, Damer et al. \cite{damer2023mordiff} proposed a diffusion-based approach, which encodes images into semantic and stochastic latent subspaces, performs linear interpolation in the semantic space and spherical linear interpolation in the stochastic space, and reconstructs morphs with improved fidelity and realism.

FLOWING \cite{bizzi2025flowing} uses a flow-based formulation that models morphing as a differential vector field, inherently ensuring continuity, invertibility, and structural preservation. By embedding flow properties into the network, it achieves accurate morphing of both 2D images and 3D shapes. LADIMO \cite{10744444} trains a diffusion model conditioned on face recognition system (FRS) latent representations: morphs are generated by blending the latent representations of contributing faces and then inverting the merged template to the image space.
\subsection{Morph Attack Detection}
To address the threat posed by face morphs, several Morph Attack Detection techniques have been proposed \cite{caldeira2023unveiling,damer2021pw,fang2022unsupervised,huber2022syn,neto2022orthomad,ramachandra2019detecting}. Based on the operating criteria, Morph Attack Detection is categorized into two approaches: 

\subsubsection{Single Image-Based Morphing Attack Detection (S-MAD)}
S-MAD methods aim to detect face morphing attacks using only a single input image. These approaches can be broadly categorized based on the type of features they exploit: texture, image quality, residual noise, deep learning, and hybrid strategies.  

Texture-based methods leverage the subtle differences between real and morphed facial images. For example, Venkatesh et al. \cite{9190629} extracted high-frequency components from multiple color spaces and computed texture descriptors, including Local Binary Patterns (LBP), Histogram of Gradients (HOG), and Binary Statistical Image Features (BSIF), to assign a morphing score. Quality-based approaches quantify degradation in image properties; Makrushin et al. \cite{DBLP:conf/visapp/MakrushinND17} proposed analyzing JPEG compression by extracting Benford features from quantized DCT coefficients to identify morphs. Zhang et al.~\cite{8486607} approached the problem from an image source perspective, treating morphed images as computer-generated and real images as natural, and exploited differences in sensor pattern noise (SPN) in specific frequency bands for detection.  

Residual noise-based methods focus on artifacts introduced during morphing. Venkatesh et al.~\cite{9190629} introduced a multi-scale context aggregation network (MS-CAN) to denoise the input image and produce a residual noise map indicative of morphing. Deep learning-based approaches extract features directly from images; for instance, Ramachandra et al.~\cite{8014962} fused features obtained from pre-trained AlexNet and VGG18 networks, while Zhang et al.~\cite{10.1007/978-3-030-95398-0_2} proposed a multi-scale attention CNN that recursively identifies morphing artifacts, achieving precise localization and strong detection performance. Hybrid S-MAD methods combine handcrafted features, morphing scores, and classifier outputs~\cite{9011378,ramachandra2019detecting,8778488} to build robust morph detection models.  

Recently, Neto et al.~\cite{neto2022orthomad} assessed whether a given sample contained two distinct identities by disentangling its identity information into two orthogonal latent vectors. In a related approach, Caldeira et al. \cite{caldeira2023unveiling} trained an autoencoder on bonafide samples to transfer identity knowledge to a morph attack detection  system, employing separate distillation strategies for bonafide and morphed images. ViT-based architectures have also been applied to MAD, achieving promising performance~\cite{zhang2024generalized}. Additionally, \cite{DBLP:conf/iwbf/IvanovskaS23} proposed a self-supervised diffusion model that reconstructs bonafide images from noisy inputs; since the model is trained exclusively on genuine samples, morphed inputs yield higher reconstruction errors, which can be exploited for anomaly-based detection. More recently, MADation \cite{DBLP:conf/wacv/CaldeiraOCIPBSD25} finetunes a foundation model along with a lightweight classifier head to perform MAD.

\subsubsection{Differential Image-Based Morphing Attack Detection (D-MAD)}
D-MAD methods exploit a reference image of the genuine subject, typically captured in a trusted environment, to determine whether a given image is morphed. Feature difference-based methods extract feature vectors from both the reference and suspect images and measure discrepancies to detect morphing. Examples include landmark shift analysis~\cite{damer2019detecting,10.1007/978-3-319-94211-7_48 } and deep face representation comparisons~\cite{9067912}. Recently, demorphing based approaches \cite{9091520,ref51} have also been proposed to perform MAD.

\subsection{Diffusion Models}
\label{sec:diffusion_prelim}

Diffusion models~\cite{NEURIPS2020_4c5bcfec} are probabilistic generative models that define a Markov chain in which samples are generated by gradually denoising a normally distributed seed. They consist of a fixed \emph{forward diffusion process} that incrementally adds Gaussian noise to data, and a learned \emph{reverse denoising process} parameterized by a neural network $\boldsymbol{\epsilon}_\theta$.

\textbf{\textit{Forward diffusion process:}}
Given a data sample $\mathbf{x}_0 \sim q(\mathbf{x}_0)$, the forward process constructs a sequence of latent variables $\{\mathbf{x}_t\}_{t=1}^T$ by progressively adding Gaussian noise:
\begin{equation}
q(\mathbf{x}_t \mid \mathbf{x}_{t-1}) =
\mathcal{N}\big(\mathbf{x}_t; \sqrt{1-\beta_t}\mathbf{x}_{t-1}, \beta_t \mathbf{I}\big),
\end{equation}
where $\{\beta_t\}_{t=1}^T$ is a predefined variance schedule. This process admits a closed-form expression:
\begin{equation}
q(\mathbf{x}_t \mid \mathbf{x}_0) =
\mathcal{N}\big(\mathbf{x}_t; \sqrt{\bar{\alpha}_t}\mathbf{x}_0, (1-\bar{\alpha}_t)\mathbf{I}\big),
\end{equation}
with $\alpha_t = 1 - \beta_t$ and $\bar{\alpha}_t = \prod_{s=1}^t \alpha_s$.

\textbf{\textit{Reverse denoising process:}}
The generative model learns a parameterized reverse Markov chain
\begin{equation}
p_\theta(\mathbf{x}_{t-1} \mid \mathbf{x}_t) =
\mathcal{N}\big(\mathbf{x}_{t-1}; \boldsymbol{\mu}_\theta(\mathbf{x}_t, t), \boldsymbol{\Sigma}_t\big),
\end{equation}
which iteratively denoises a sample starting from $\mathbf{x}_T \sim \mathcal{N}(\mathbf{0}, \mathbf{I})$.
In practice, the model is trained to predict the injected noise $\boldsymbol{\epsilon}$ using the objective
\begin{equation}
\mathcal{L}_{\mathrm{DM}} =
\mathbb{E}_{\mathbf{x}_0, \boldsymbol{\epsilon}, t}
\left[
\left\|
\boldsymbol{\epsilon} -
\boldsymbol{\epsilon}_\theta
\left(
\sqrt{\bar{\alpha}_t}\mathbf{x}_0 +
\sqrt{1-\bar{\alpha}_t}\boldsymbol{\epsilon}, t
\right)
\right\|_2^2
\right].
\end{equation}

\textbf{\textit{Text-conditioned diffusion:}}
Text-to-image (T2I) diffusion models extend the denoising process to be conditioned on textual descriptions. Given a text prompt, a pretrained language encoder (e.g., CLIP~\cite{radford2021learning} or T5~\cite{10.5555/3455716.3455856}) produces a sequence of embeddings $\mathbf{c} = \{\mathbf{c}_i\}_{i=1}^N$. The denoising network $\boldsymbol{\epsilon}_\theta(\mathbf{x}_t, t, \mathbf{c})$ incorporates these text embeddings via cross-attention mechanisms, where spatial features query the text representations to modulate the denoising predictions. To improve sample quality and adherence to the prompt, classifier-free guidance~\cite{ho2021classifierfree} is commonly employed during inference. 

\textbf{\textit{Latent diffusion models:}}
Operating directly in pixel space is computationally expensive for high-resolution images. Latent diffusion models (LDMs)~\cite{Rombach2021HighResolutionIS} address this by applying diffusion in the latent space of a pretrained autoencoder. An encoder $\mathcal{E}$ compresses an image $\mathbf{x}_0 \in \mathbb{R}^{H \times W \times 3}$ into a lower-dimensional latent representation
\begin{equation}
\mathbf{z}_0 = \mathcal{E}(\mathbf{x}_0) \in \mathbb{R}^{h \times w \times c},
\end{equation}
where typically $h = H/f$, $w = W/f$ with downsampling factor $f \in \{4, 8, 16\}$. The forward and reverse diffusion processes then operate on $\mathbf{z}_0$ rather than $\mathbf{x}_0$. After denoising, a decoder $\mathcal{D}$ reconstructs the image: $\hat{\mathbf{x}}_0 = \mathcal{D}(\hat{\mathbf{z}}_0)$. This architecture, employed by Stable Diffusion~\cite{Rombach2021HighResolutionIS}, dramatically reduces computational requirements while maintaining high perceptual quality, enabling efficient training and inference for megapixel-scale image synthesis.

\subsection{Multimodal Large Language Models (MLLM)}
Multimodal large language models (MLLMs) are multimodal architectures designed to generate textual outputs conditioned on both visual and text inputs. Recent advances in large-scale MLLMs have demonstrated strong zero-shot and generalization capabilities across diverse visual domains, including natural images, documents, and web content \cite{10.5555/3666122.3667638,instructblip,Achiam2023GPT4TR,claude,gemini}. Architecturally, most MLLMs comprise three core components: (i) a vision encoder, such as CLIP \cite{radford2021learning} or SigLIP \cite{10377550} to extract visual representations, (ii) a large language model responsible for text understanding and generation, and (iii) a projection module that aligns visual features with the language embedding space. These models are typically trained using large-scale image–caption pairs and instruction-tuning datasets. Beyond pretraining, recent post-training strategies have further enhanced MLLM capabilities, particularly in conversational grounding \cite{xiong2024llavaovchat} and multi-step reasoning \cite{wang2024enhancing}.

\begin{table*}[t]
\centering
\caption{Summary of existing facial demorphing methods. The upper section presents 
reference-based approaches, whereas the lower section summarizes reference-free methods.}
\label{tab:demorph_review}
\setlength{\extrarowheight}{3pt}
\resizebox{\textwidth}{!}{%
\begin{tabular}{@{} l l l p{5cm} p{5cm} @{}}
\toprule
\rowcolor{gray!20}
\textbf{Method} & \textbf{Core Approach} & \textbf{Domain} & \textbf{Key Limitations} & \textbf{Metric} \\
\midrule

\multicolumn{5}{@{}l@{}}{\textit{\textbf{\small Reference-Based Methods}}} \\[2pt]
\midrule

Ferrara~\cite{ferrara2017face}
& Heuristic-based morph inversion
& Pixel space
& Limited generalization capability; non-learned formulation
& Genuine Acceptance Rate \\

Peng ~\cite{8730323}
& GAN-based conditional reconstruction
& Pixel space
& Sensitive to morph artifacts and identity imbalance
& Restoration Accuracy \\

Ortega-Delcampo ~\cite{ortega2020border}
& Autoencoder within MAD pipeline
& Pixel space
& Limited identity disentanglement capacity
& BPCER, APCER \\

Long ~\cite{10415238}
& Diffusion autoencoder with dual identity branches
& Latent space
& High computational cost; relies on latent separability assumptions
& Restoration Accuracy \\

Zhang ~\cite{zhang2023morphganformer}
& Transformer-based GAN with latent decomposition
& Latent space
& Inconsistent identity preservation under complex morphs
& Restoration Accuracy \\

Cai ~\cite{Cai2025ASF}
& StyleGAN $\mathcal{W}^+$ separation with cross-attention
& Latent space
& Performance constrained by inversion fidelity
& Restoration Accuracy \\

\midrule
\multicolumn{5}{@{}l@{}}{\textit{\textbf{\small Reference-Free Methods}}} \\[2pt]
\midrule

Banerjee ~\cite{ref51}
& GAN with multi-discriminator separation
& Pixel space
& Residual morph artifacts; weak disentanglement robustness
& TMR @ FMR \\

Shukla~\cite{ref18}
& Diffusion-based reverse denoising reconstruction
& Pixel space
& Assumes identity overlap between training and testing sets
& Restoration Accuracy \\

Shukla ~\cite{ipd}
& Privacy-preserving component decomposition
& Pixel space
& Requires shared identities across data splits
& Restoration Accuracy \\

Shukla ~\cite{lcgan}
& Latent splitting in pretrained autoencoder space
& Latent space
& Sensitive to encoder inversion fidelity
& Restoration Accuracy, SSIM, PSNR, TMR \\

Shukla ~\cite{dcgan}
& Dual conditioning (RGB + CLIP latent embeddings)
& Hybrid (Pixel + Latent)
& Limited robustness to advanced generative morphs
& Restoration Accuracy, SSIM, PSNR, TMR \\

Shukla ~\cite{diffdemorph}
& Diffusion-based coupled one-to-one image translation
& Pixel space
& Performance degrades on high-quality generative morphs
& Restoration Accuracy, BW-IQA~\cite{shukla2025metric} \\

\bottomrule
\end{tabular}%
}
\end{table*}
\section{Related Work}
\subsection{Reference-Based Demorphing}
In reference-based facial demorphing, the algorithm is provided with both the morph image and a trusted reference image, typically a live capture of one of the contributors. The objective is to recover the accomplice’s facial image by leveraging identity information from the reference sample.

Ferrara \etal~\cite{ferrara2017face} first demonstrated successful restoration of an accomplice’s face by reversing the morphing process. Peng \etal~\cite{8730323} extended this direction using generative adversarial networks (GANs) to reconstruct the accomplice’s image from a morph–reference pair. Ortega-Delcampo \etal~\cite{ortega2020border} employed an autoencoder-based framework that restores facial images within a morph attack detection pipeline. More recently, Long \etal~\cite{10415238} proposed a diffusion-based framework built upon pre-trained diffusion autoencoders. They introduced a dual-branch identity separation network operating in semantic latent space to disentangle the latent representations of the two contributors. Zhang \etal~\cite{zhang2023morphganformer} presented a transformer-based GAN (MorphGANFormer) and formulated demorphing as latent decomposition, modeling the morph representation as $M = G_1 + G_2$ where $M,G_1$ and $G_2$ are morph and constituent images respectively. Demorphing is achieved by subtracting the latent code of the trusted reference from the morph latent representation, followed by image synthesis from the residual code.

Cai \etal~\cite{Cai2025ASF} utilized a StyleGAN generator for high-fidelity synthesis. A pre-trained encoder extracts multi-scale feature maps (at $\frac{1}{2}$, $\frac{1}{4}$, and $\frac{1}{8}$ resolutions) to generate latent codes in the $\mathcal{W}^+$ space of the StyleGAN for both morph and reference image. These codes are concatenated and passed through a feature separation module composed of multi-scale convolutional kernels and a cross-attention mechanism to disentangle identity information. The refined latent code is then used by the StyleGAN generator to reconstruct the accomplice’s face.

\subsection{Reference-Free Demorphing}

In the reference-free (RF) demorphing setting, only the morph image is available at inference time, rendering the problem significantly more challenging than its differential counterpart. The objective is to recover two plausible and identity-consistent constituent faces from a single morphed observation, without access to auxiliary identity cues.

Early work by Banerjee \etal~\cite{ref51} introduced a GAN-based framework consisting of a generator and three discriminators for image separation. While effective for natural scene decomposition originally, the method encountered a \emph{morph-replication} artifact in the demorphing, where both generated outputs closely resembled the input morph rather than disentangling distinct identities. 

Shukla~\cite{ref18} proposed a diffusion-based demorphing approach that progressively adds noise to the morph and reconstructs constituent faces during the reverse diffusion process. However, this formulation assumes that both training and testing morphs are generated from the same set of identities, thereby limiting its generalization to open-set scenarios. Similarly, Shukla  and Ross~\cite{ipd} decomposed the morph into multiple unintelligible, privacy-preserving components and reconstructed the constituent faces through learned weighted combinations. This approach also presumes shared identity distributions across training and testing data. Latent-domain strategies have also been explored. In~\cite{lcgan}, a GAN-based architecture operates in the latent space of a pre-trained autoencoder, attempting to split the morph latent representation into constituent identity latents. dc-GAN~\cite{dcgan} further introduced a dual-conditioned framework leveraging both the RGB morph and CLIP-based latent embeddings to improve disentanglement. More recently, Shukla and Ross~\cite{diffdemorph} presented a diffusion-based one-to-one image translation framework to jointly disentangle morphs into their constituent images.

Despite these advances, existing RF demorphing methods remain constrained by assumptions on identity overlap or morph generation mechanisms. In particular, maintaining high identity fidelity while ensuring photorealistic reconstruction remains challenging when presented with modern GAN and diffusion-generated morphs, highlighting the need for more robust and generalizable demorphing strategies. An overview of these methods is presented in Table \ref{tab:demorph_review}.

\section{Proposed Method}

We formulate facial demorphing as a conditional generation problem aimed at 
recovering both constituent face images from a single morph image. Our approach 
treats demorphing as a coupled one-to-one generation process, where the two faces 
are synthesized jointly rather than independently. This joint formulation ensures 
that the generation of each face is interdependent, helping to alleviate morph 
replication while preserving consistency with the original constituent identities.

Let $x$ denote a morph image generated from two face images $i_1$ and $i_2$. In 
the RGB domain, the constituent images exist independently, while the morph image 
establishes a relationship between them through the triplet $(x, i_1, i_2)$. We 
therefore represent the demorphing target as a paired entity $i = (i_1, i_2)$ and 
aim to model the conditional distribution $p(i \mid x)$. To this end, we propose 
a coupled Denoising Diffusion Probabilistic Model (DDPM) 
\cite{NEURIPS2020_4c5bcfec}, augmented with multimodal semantic guidance extracted 
from a Multimodal Large Language Model (MLLM).

\begin{figure*}
    \centering
    \includegraphics[width=0.95\linewidth]{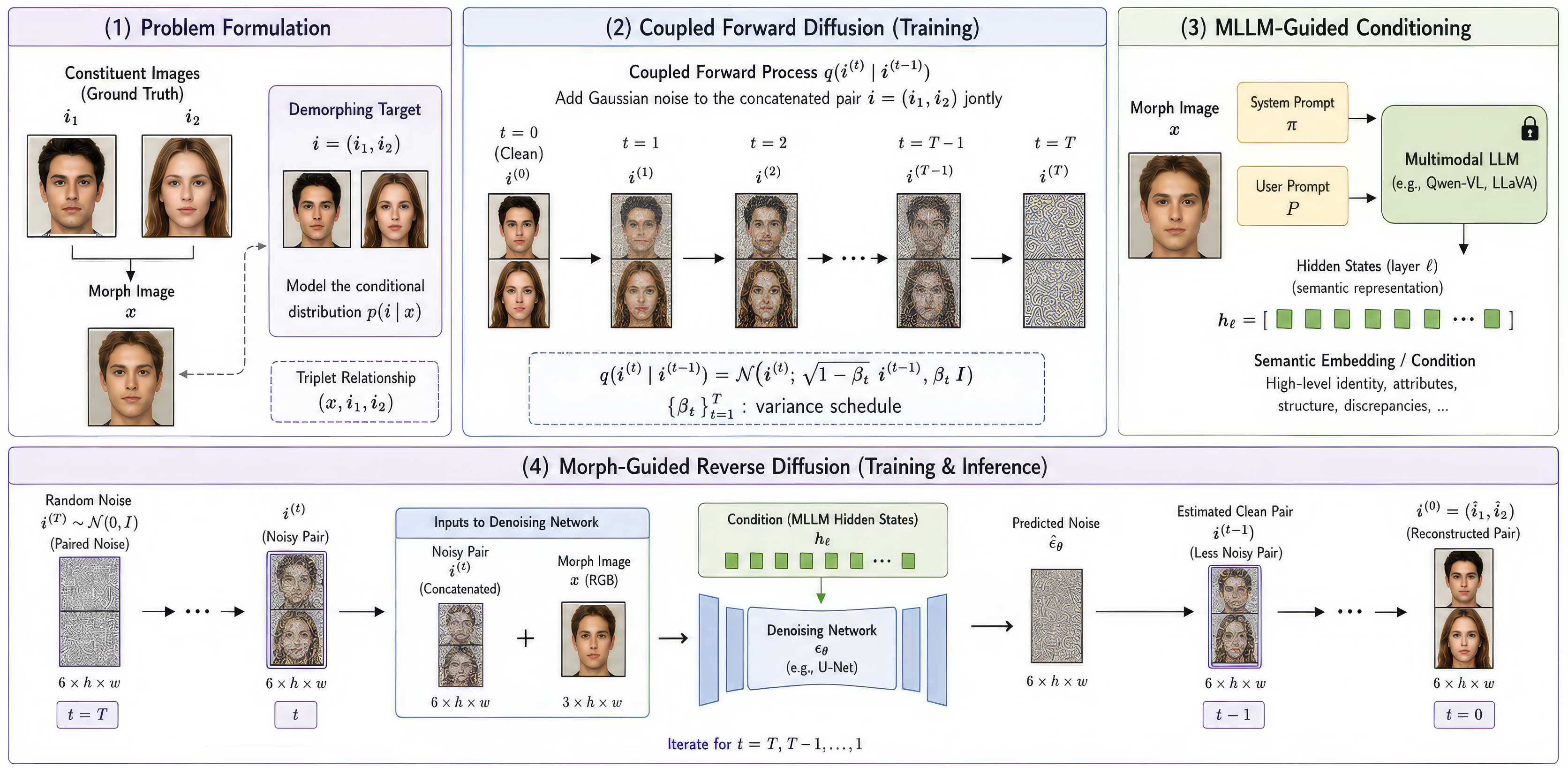}
    \caption{\textbf{Overview of the proposed prompt-driven demorphing method.} 
    A morphed face image is provided to a Multimodal Large Language Model (MLLM) 
    to generate a semantic description, from which hidden states are extracted from 
    intermediate transformer layers. These hidden representations condition a 
    diffusion-based image-to-image model, which takes the morphed image as input 
    and reconstructs the constituent source images. See 
    Section~\ref{sec:ablation} for detailed prompts used in this study.}
    \label{fig:main}
\end{figure*}

\subsection{Coupled Forward Diffusion}

The coupled forward diffusion process incrementally corrupts the paired 
ground-truth images $i = (i_1, i_2) \in \mathbb{R}^{6\times H\times W}$ by 
adding Gaussian noise over $T$ diffusion steps. The joint marginal at step $t$ 
is given in closed form by:
\begin{equation}
    q\!\left(i^{(t)} \mid i^{(0)}\right) 
    = \mathcal{N}\!\left(i^{(t)};\; 
      \sqrt{\bar{\alpha}_t}\, i^{(0)},\; 
      (1 - \bar{\alpha}_t)\,\mathbf{I}\right),
\end{equation}
where, $\bar{\alpha}_t = \prod_{s=1}^{t}(1 - \beta_s)$, 
$\{\beta_t \in (0,1)\}_{t=1}^T$ is a predefined variance schedule, and 
$\mathbf{I}$ denotes the  covariance operator over the joint 
6-channel space $\mathbb{R}^{6 \times H \times W}$. As $T \to \infty$, 
$i^{(T)}$ converges to an isotropic Gaussian distribution.

Crucially, noise is applied in the \emph{joint} space 
$\mathbb{R}^{6\times H\times W} $ rather 
than independently to each constituent image. This preserves the cross-image 
dependency structure throughout the forward process: because both outputs share 
a single noise trajectory, the reverse process must simultaneously denoise both 
channels, preventing the network from collapsing to independent marginal 
denoising of $i_1$ and $i_2$ separately.

\subsection{MLLM-Guided Conditioning}

A key departure from prior diffusion-based demorphing approaches lies in how 
conditioning information is obtained and injected. Given a morph image $x$, we 
first pass it to a Multimodal Large Language Model (MLLM) along with a system 
prompt $\pi$ and a user prompt $P$. The prompts are designed to elicit detailed 
semantic reasoning about the morph, such as facial attributes, identity cues, 
and visual discrepancies indicative of multiple contributors.

\textbf{\textit{Hidden state extraction:}}
Rather than using the MLLM's decoded text output, we extract the hidden state 
representation from an intermediate transformer layer $\ell$:
\begin{equation}
    h_\ell = f_\ell(x, \pi, P) \in \mathbb{R}^{N \times d},
\end{equation}
where, $N$ is the number of output tokens and $d$ is the hidden dimension at 
layer $\ell$. These representations are projected via a learned linear map 
$W_{\mathrm{proj}} \in \mathbb{R}^{d \times d_{\mathrm{cross}}}$ to match the 
cross-attention dimension $d_{\mathrm{cross}}$ expected by the denoising 
network, yielding the conditioning tensor 
$\tilde{h}_\ell = h_\ell W_{\mathrm{proj}}^\top \in 
\mathbb{R}^{N \times d_{\mathrm{cross}}}$.
This approach bypasses the lossy cycle of decompressing the MLLM's internal 
knowledge into discrete text tokens and re-encoding it with a separate text 
encoder, instead directly exposing the model's rich multimodal reasoning to the 
diffusion process.

\textbf{\textit{Conditioning mechanism:}}
We adopt a \texttt{UNet2DConditionModel} \cite{von-platen-etal-2022-diffusers} 
as the denoising backbone.  At each 
transformer block inside the UNet, the intermediate spatial feature map 
$z_t \in \mathbb{R}^{(H'W') \times d_{\mathrm{model}}}$ attends to the 
conditioning sequence $\tilde{h}_\ell$ through a built-in cross-attention layer:
\begin{equation}
    \mathrm{CrossAttn}(z_t,\, \tilde{h}_\ell) 
    = \mathrm{softmax}\!\left(
        \frac{(W_Q\, z_t)\,(W_K\, \tilde{h}_\ell)^\top}{\sqrt{d_k}}
      \right)(W_V\, \tilde{h}_\ell),
\end{equation}
where, $W_Q \in \mathbb{R}^{d_k \times d_{\mathrm{model}}}$ and 
$W_K, W_V \in \mathbb{R}^{d_k \times d_{\mathrm{cross}}}$ are the learned 
query, key, and value projection matrices respectively, and $d_k$ is the 
attention head dimension. 

This injects high-level semantic cues about facial identity, gender, age, and 
structure at every scale of the UNet, complementing the low-level visual 
information carried by the morph image $x$ in the RGB domain, enabling generation informed by both 
appearance and meaning.

\subsection{Morph-Guided Reverse Diffusion}

The learning objective is to estimate the conditional distribution 
$p((i_1, i_2) \mid x)$. Following the DDPM framework 
\cite{NEURIPS2020_4c5bcfec}, the reverse process is defined by:
\begin{equation}
    p_\theta\!\left(i^{(t-1)} \mid i^{(t)}, x, h_\ell\right) 
    = \mathcal{N}\!\left(i^{(t-1)};\; 
      \mu_\theta\!\left(i^{(t)}, x, h_\ell, t\right),\; 
      \sigma_t^2\,\mathbf{I}\right).
\end{equation}
The training objective aims to  reduce the squared 
error between the true and predicted noise:
\begin{equation}
    \mathcal{L} 
    = \mathbb{E}_{x,\; i \sim p(i\mid x),\; 
                  \epsilon \sim \mathcal{N}(0,\mathbf{I}),\; t}
      \!\left[
        \left\lVert 
          \epsilon 
          - \epsilon_\theta\!\left(i^{(t)},\, x,\, \tilde{h}_\ell,\, t\right) 
        \right\rVert_2^2
      \right],
\end{equation}
where, $i^{(t)}$ denotes the noisy coupled sample at time step $t$, $\epsilon$ 
is the sampled Gaussian noise, and $\tilde{h}_\ell$ is the  MLLM 
hidden state passed to the denoising 
network. Providing the morph image $x$ explicitly in the RGB domain alongside 
the semantic conditioning allows the denoising network to attend to fine-grained 
perceptual details---such as hair, background, and subtle facial cues---that are 
often lost in heavily compressed latent representations.

\textbf{\textit{Inference:}}
Starting from $o^{(T)} \sim \mathcal{N}(0, \mathbf{I})$, the morph image $x$ 
is concatenated channel-wise with the noisy sample at each timestep to form the 
9-channel input $[o^{(t)};\, x] \in \mathbb{R}^{9 \times H \times W}$, and the 
reverse chain iterates the following update for $t = T, T{-}1, \ldots, 1$:
\begin{equation}
    o^{(t-1)} 
    = \frac{1}{\sqrt{\alpha_t}}
      \left(
        o^{(t)} 
        - \frac{1 - \alpha_t}{\sqrt{1 - \bar{\alpha}_t}}\,
          \epsilon_\theta\!\left(
            [o^{(t)};\, x],\, \tilde{h}_\ell,\, t
          \right)
      \right)
      + \sigma_t\, z
\end{equation}
%
until $t = 0$, yielding the final reconstructed constituent image pair 
$(o_1, o_2) = \mathrm{split}(o^{(0)})$, where 
$\mathrm{split}(\cdot)$ partitions the 6-channel output along the channel 
dimension into two RGB images.  A schematic overview of the proposed architecture is shown 
in Figure~\ref{fig:main}.

\subsection{Solution Space Constraints}

The reference-free demorphing problem is fundamentally ill-posed: given a morph
$x = \mathcal{M}(i_1, i_2)$, infinitely many pairs $(i_1', i_2')$ satisfy
$\mathcal{M}(i_1', i_2') \approx x$, including degenerate solutions such as
$(x, x)$. Our formulation restricts the effective solution space through three 
complementary mechanisms.

\textbf{(i) Data-supported prior.}
Training exclusively on ground-truth constituent pairs biases score estimation 
toward the data-supported joint manifold $p_{\mathrm{data}}(i_1, i_2 \mid x)$, 
eliminating decompositions that produce unnatural or out-of-distribution faces 
and restricting solutions to the support of realistic identity pairs.

\textbf{(ii) Joint denoising dependency.}
The 6-channel coupled denoising enforces inter-output co-dependency: because 
both outputs share a single noise trajectory and network, 
$\mathrm{Cov}(\hat{i}_1, \hat{i}_2 \mid x) \neq 0$ is implicitly encouraged, 
discouraging independent or near-identical reconstructions and penalising 
degenerate solutions such as $(x, x)$.

\textbf{(iii) MLLM semantic prior.}
The hidden state conditioning injects identity-level semantic constraints, 
soft-constraining the output configuration via the factorisation:
\begin{equation}
    p\!\left((i_1, i_2) \mid x, \tilde{h}_\ell\right) 
    \propto p(x \mid i_1, i_2)\cdot p(i_1, i_2 \mid \tilde{h}_\ell).
\end{equation}

We note that these constraints are necessary but not sufficient to guarantee a 
unique decomposition. DDPM training minimises expected $\ell_2$ noise-prediction 
error, which under a multimodal posterior favours the posterior mean rather than 
a specific mode, and formal guarantees on output diversity remain an open 
question.

\section{Experimental Setup}

\subsection{Datasets}
A major challenge in reference-free face demorphing is the scarcity of large-scale morph datasets, primarily due to privacy and ethical constraints associated with biometric data. Most publicly available face morph datasets are designed for Morph Attack Detection (MAD) and typically contain on the order of $10^3$ morphs, which is insufficient for training data-intensive generative models. To mitigate this limitation, we train our model exclusively on morphs generated from synthetically created face images. This strategy not only enables large-scale training but also alleviates privacy concerns by avoiding the use of real biometric identities. To evaluate generalization and practical applicability, we test the proposed method on multiple real-world morph datasets.

\textbf{Training Dataset:}
During training, we construct morph images on-the-fly by randomly sampling two bonafide face images from the SMDD dataset~\cite{SMDD}. Since the original SMDD morphs are tailored for MAD and rely on a limited set of base images, making them unsuitable for demorphing, we instead generate new morphs dynamically. Morph generation is performed using the widely adopted OpenCV/dlib-based morphing pipeline~\cite{10.5555/1577069.1755843,openCVmorph}, which leverages Dlib’s facial landmark detector~\cite{Lee1980}. Using this procedure, we generate 15,000 training morphs from the SMDD training split.

All images are processed using MTCNN~\cite{7553523} for face detection, after which only the detected facial regions are cropped. The cropped faces are resized to $256 \times 256$ and normalized. Samples for which face detection fails are discarded. No additional geometric augmentations are applied, ensuring that key facial structures (e.g., eyes, nose, and mouth) remain spatially aligned between the morphs and their corresponding ground-truth constituent images.

\textbf{Testing Dataset:}
We evaluate the proposed method on three widely used face morph datasets: AMSL~\cite{ref64}, FRLL-Morphs~\cite{DeBruine2021}, and MorDiff~\cite{damer2023mordiff}. The FRLL-Morphs dataset includes morphs generated using four different techniques: OpenCV~\cite{openCVmorph}, StyleGAN~\cite{stylegan}, WebMorph~\cite{debruine2018debruine}, and FaceMorph~\cite{quek2019facemorpher}. Across all datasets, the source bonafide images are drawn from the FRLL dataset, which contains 102 bonafide identities, each represented by two frontal images (neutral and smiling), yielding a total of 204 source images.

The number of morphs in each dataset is as follows: AMSL—2,175; FaceMorph—1,222; StyleGAN—1,222; OpenCV—1,221; WebMorph—1,221; and MorDiff—1,000. Collectively, these datasets cover both traditional landmark-based morphing methods and modern generative approaches, enabling a comprehensive evaluation of demorphing performance. Detailed dataset statistics are presented in Table \ref{tab:datasets}.

\begin{table}[htbp]
\centering
\caption{Summary of training and testing datasets used.}
\label{tab:datasets}
\resizebox{\linewidth}{!}{%
\begin{tabular}{l l l c c c}
\toprule
\textbf{Dataset} & \textbf{Split / Type} & \thead{\textbf{Morphing} \\ \textbf{Technique}} & \textbf{\# Identities} & \textbf{\# Morphs} & \thead{\textbf{Landmark} \\ \textbf{-based}} \\
\midrule
SMDD~\cite{SMDD} & Training & OpenCV/dlib  &  \textemdash & 15,000 & \cmark \\
\midrule
AMSL~\cite{ref64} & Test & Landmark-based & 102 & 2,175 & \cmark \\
FRLL~\cite{DeBruine2021} & Test & OpenCV  & 102 & 1,221 & \cmark \\
FRLL~\cite{DeBruine2021} & Test & StyleGAN & 102 & 1,222 & \xmark \\
FRLL~\cite{DeBruine2021} & Test & WebMorph  & 102 & 1,221 & \cmark \\
FRLL~\cite{DeBruine2021} & Test & FaceMorph  & 102 & 1,222 & \cmark \\
MorDiff~\cite{damer2023mordiff} & Test & Diffusion  & 102 & 1,000 & \xmark \\
\bottomrule
\end{tabular}%
}
\end{table}

\subsection{MLLM-Based Conditioning Embeddings}

To guide the demorphing process, we leverage pretrained Multimodal Large Language Models (MLLMs) to extract semantically rich visual embeddings from morphed face images. Specifically, we employ Qwen3-VL-8B \cite{qwen3} and LLaVA-1.6-mistral-7B \cite{liu2023improved} in this work. We investigate the impact of embeddings extracted from different depths of the MLLM, including \textit{initial}, \textit{middle}, and \textit{last} (Layer 1, 18 and 36 for Qwen and 1,16 and 32 for LLaVA) transformer blocks, as well as the \textit{ViT-level} representation. All embeddings are precomputed offline and stored for efficient training. Since the embeddings have variable sequence lengths, we apply zero-padding within each batch while maintaining a fixed embedding dimensionality to match the cross-attention projection layers of the diffusion network. Separate models are trained for each embedding type to analyze how different levels of MLLM abstraction influence demorphing performance.

\subsection{Conditional Diffusion Model}
We adopt a conditional DDPM framework with a UNet backbone operating directly in the image space. The UNet processes images at a spatial resolution of $256 \times 256$ and receives a 9-channel input formed by concatenating the noisy target image pair (6 channels) with the morphed image (3 channels). At each diffusion timestep, the network is trained to predict the Gaussian noise added to the two constituent images.

The UNet architecture comprises six downsampling and six upsampling blocks with channel dimensions $(128, 128, 256, 256, 512, 512)$, and incorporates self-attention modules at intermediate spatial resolutions to capture long-range dependencies. Conditioning information derived from the MLLM embeddings is injected via cross-attention layers at multiple stages of the UNet, allowing the denoising process to exploit high-level semantic and identity-related cues in conjunction with pixel-level guidance from the morph image. 

\subsection{Training Details}

We train the model using a DDPM noise scheduler with 1000 diffusion timesteps. At each training iteration, Gaussian noise is added to the concatenated ground-truth image pair according to a randomly sampled timestep, and the network is optimized to predict the injected noise using a mean squared error (MSE) loss.

Training is performed for 500 epochs with a batch size of 16 and gradient accumulation over 2 steps. We use the Adam \cite{kingma2017adammethodstochasticoptimization} optimizer with a learning rate of $1\times10^{-4}$ and employ a cosine learning rate schedule with 500 warm-up steps. Mixed-precision (FP16) training and multi-GPU acceleration are enabled using the \texttt{Accelerate} \cite{accelerate} library.

\subsection{Evaluation Protocol}

To quantitatively evaluate demorphing performance, we measure identity similarity between the generated images and their corresponding ground-truth faces using AdaFace~\cite{kim2022adaface} and ArcFace~\cite{deng2019arcface}. We report \emph{Restoration Accuracy}, defined as the fraction of reconstructed identities correctly matched to their constituent ground-truth faces at  False Match Rate (FMR) thresholds of 0.1\%, 1\%, and 10\%. 

As demorphing produces two outputs per morph without a predefined order, we adopt a permutation-invariant matching strategy. Given ground-truth images $(\mathrm{GT}_1, \mathrm{GT}_2)$ and outputs $(\mathrm{OUT}_1, \mathrm{OUT}_2)$, we evaluate both assignments:
$(\mathrm{GT}_1 \leftrightarrow \mathrm{OUT}_1,\; \mathrm{GT}_2 \leftrightarrow \mathrm{OUT}_2)$ and
$(\mathrm{GT}_1 \leftrightarrow \mathrm{OUT}_2,\; \mathrm{GT}_2 \leftrightarrow \mathrm{OUT}_1)$.
The pairing with the higher sum of similarities is selected, and accuracy is computed based on this optimal assignment.

To assess the visual fidelity of the generated images, we additionally compute standard image quality metrics, including Structural Similarity Index Measure (SSIM) and Peak Signal-to-Noise Ratio (PSNR).

\subsection{Implementation Details}

All experiments are implemented in PyTorch using the \texttt{Diffusers} \cite{von-platen-etal-2022-diffusers} and \texttt{Transformers} \cite{wolf-etal-2020-transformers} libraries. MLLM embeddings are extracted using a pretrained vision-language model and cached for efficiency. Identity embeddings are obtained using pretrained AdaFace and ArcFace models. All hyperparameters are kept fixed across different embedding variants to ensure controlled and fair comparisons.

\begin{figure}
    \centering
    \includegraphics[width=1\linewidth]{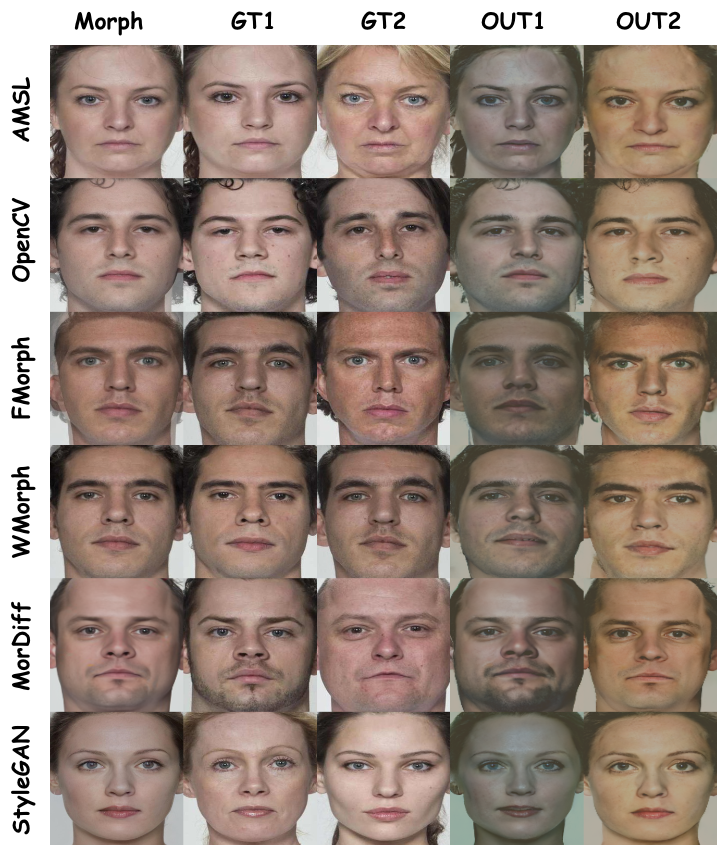}
    \caption{Reconstruction results of our method on samples from six different morphing techniques. Each morph is generated using two ground-truth identities, denoted as GT1 and GT2. Our method reconstructs the constituent identities from the morph, producing the corresponding outputs OUT1 and OUT2.}
    \label{fig:main-result2}
\end{figure}
\section{Results and Discussion}
\begin{table*}[ht]
\centering
\caption{Comparison of Restoration Accuracy across different demorphing techniques under a unified protocol. Our method outperforms existing baselines by a considerable margin. Early approaches relied on fixed similarity thresholds (0.4), which corresponds to ${>}10\%$ FMR. At $0.1\%$ FMR, \textbf{\textcolor{cvprgold}{first}} and \textcolor{cvprblue}{second} best results per column are highlighted in \textbf{\textcolor{cvprgold}{orange}} and \textcolor{cvprblue}{green}, respectively. We use middle-layer embeddings as guiding signal  for both MLLMs.}
\label{tab:restoration_accuracy}
\resizebox{\linewidth}{!}{
\begin{tabular}{@{}c|c|cccccc|cccccc@{}}
\toprule
\multirow{2}{*}{\textbf{Method}} & \multirow{2}{*}{\textbf{RA}} & \multicolumn{6}{c|}{\textbf{ArcFace}} & \multicolumn{6}{c}{\textbf{AdaFace}} \\
\cmidrule(lr){3-8} \cmidrule(lr){9-14}
& & AMSL & OpenCV & FMorph & MorDiff & Wmorph & StyleGAN & AMSL & OpenCV & FMorph & MorDiff & Wmorph & StyleGAN \\
\midrule
\multirow{3}{*}{\makecell{\methodname + \\Qwen3-VL}}
& @ $10\%$ FMR  & 99.98\% & 100.00\% & 99.93\% & 98.53\% & 99.78\% & 67.23\% & 99.74\% & 100.00\% & 100.00\% & 99.79\% & 99.91\% & 57.28\% \\
& @ $1\%$ FMR   & 99.35\% & 99.87\% & 99.14\% & 98.53\% & 99.14\% & 32.77\% & 98.95\% & 99.96\% & 99.73\% & 98.42\% & 99.27\% & 23.86\% \\
& @ $0.1\%$ FMR  & \textbf{\textcolor{cvprgold}{96.35\%}}
  & \textbf{\textcolor{cvprgold}{98.65\%}}
  & \textbf{\textcolor{cvprgold}{96.55\%}}
  & \textcolor{cvprblue}{94.63\%}
  & \textcolor{cvprblue}{95.22\%}
  & \textbf{\textcolor{cvprgold}{9.13\%}}
  & \textbf{\textcolor{cvprgold}{96.41\%}}
  & \textbf{\textcolor{cvprgold}{98.74\%}}
  & \textbf{\textcolor{cvprgold}{98.14\%}}
  & \textbf{\textcolor{cvprgold}{95.37\%}}
  & \textcolor{cvprblue}{96.25\%}
  & \textbf{\textcolor{cvprgold}{5.34\%}} \\
\midrule
\multirow{3}{*}{\makecell{\methodname + \\LLaVA-1.6}}
& @ $10\%$ FMR & 99.83\% & 99.84\% & 99.87\% & 99.76\% & 99.72\% & 56.24\% & 99.73\% & 99.91\% & 99.82\% & 99.67\% & 99.53\% & 46.74\% \\
& @ $1\%$ FMR  & 97.14\% & 99.23\% & 98.56\% & 97.52\% & 97.48\% & 23.97\% & 97.63\% & 99.24\% & 99.31\% & 97.68\% & 97.43\% & 17.04\% \\
& @ $0.1\%$ FMR & 88.82\%
  & 94.63\%
  & 92.54\%
  & 87.71\%
  & 87.74\%
  & 6.62\%
  & 90.91\%
  & \textcolor{cvprblue}{96.82\%}
  & \textcolor{cvprblue}{94.37\%}
  & 90.73\%
  & 90.53\%
  & \textcolor{cvprblue}{2.94\%} \\
\midrule
\midrule
\multirow{3}{*}{lc-GAN \cite{lcgan}}
& @ $10\%$ FMR & $99.90\%$ & $99.83\%$ & $100.00\%$ & $100.00\%$ & $100.00\%$ & $38.76\%$ & $98.58\%$ & $99.53\%$ & $99.71\%$ & $100.00\%$ & $100.00\%$ & $50.39\%$ \\
& @ $1\%$ FMR  & $99.09\%$ & $99.30\%$ & $99.03\%$ & $100.00\%$ & $99.47\%$ & $12.57\%$ & $96.35\%$ & $98.58\%$ & $98.57\%$ & $100.00\%$ & $98.91\%$ & $20.75\%$ \\
& @ $0.1\%$ FMR
  & \textcolor{cvprblue}{$96.26\%$}
  & \textcolor{cvprblue}{$95.83\%$}
  & \textcolor{cvprblue}{$95.16\%$}
  & \textbf{\textcolor{cvprgold}{$98.63\%$}}
  & \textbf{\textcolor{cvprgold}{$96.28\%$}}
  & $2.12\%$
  & \textcolor{cvprblue}{$91.19\%$}
  & $93.69\%$
  & $93.41\%$
  & \textbf{\textcolor{cvprgold}{$98.83\%$}}
  & \textcolor{cvprblue}{$94.53\%$}
  & $4.52\%$ \\
\midrule
{diffDeMorph \cite{diffdemorph}}
& @ ${>}10\%$ FMR  & $99.49\%$ & $100.00\%$ & $100.00\%$ & $100.00\%$ & $99.82\%$ & $95.97\%$ & $51.69\%$ & $58.88\%$ & $59.79\%$ & $66.13\%$ & $40.26\%$ & $67.00\%$ \\
\midrule
{IPD \cite{ipd}}
& @ ${>}10\%$ FMR & $25.69\%$ & $40.54\%$ & $37.82\%$ & $38.12\%$ & $25.61\%$ & $16.22\%$
& $0.18\%$ & $1.89\%$ & $1.43\%$ & $3.88\%$ & $0.31\%$ & $0.00\%$ \\
\midrule
{SDeMorph \cite{ref18}}
& @ ${>}10\%$ FMR & $12.56\%$ & $15.62\%$ & $13.18\%$ & $11.67\%$ & $12.80\%$ & $0.00\%$
& $0.00\%$ & $0.00\%$ & $0.00\%$ & $0.00\%$ & $0.00\%$ & $0.00\%$ \\
\midrule
Face Demorphing \cite{ref51}
& @ ${>}10\%$ FMR & $0.45\%$ & $0.53\%$ & $0.51\%$ & $0.62\%$ & $0.50\%$ & $0.43\%$
& $0.17\%$ & $0.23\%$ & $0.17\%$ & $0.29\%$ & $0.20\%$ & $0.00\%$ \\
\bottomrule
\end{tabular}
}
\end{table*}

\begin{table}[ht]
\centering
\caption{Comparison of demorphing quality using PSNR metric. Higher values indicate better image reconstruction fidelity.}
\label{tab:psnr}
\resizebox{\linewidth}{!}{
\begin{tabular}{@{}c|cccccc@{}}
\toprule
\textbf{Method} & AMSL & OpenCV & FMorph & WMorph & MorDiff & StyleGAN \\
\midrule
\makecell{\methodname + \\Qwen3-VL} & \textbf{18.36} & \textbf{18.34} & \textbf{18.26} & \textbf{17.99} & \textbf{18.55} & \textbf{19.56} \\
\makecell{\methodname + \\LLaVA-1.6} & 17.32& 17.22 &17.34 & 17.14 & 17.35 & 18.01 \\

\midrule
lc-GAN \cite{lcgan} & 10.81 & 11.65 & 11.63 & 11.17 & 10.93 & 10.18 \\
IPD (2024) \cite{ipd} & 9.32 & 10.37 & 10.27 & 9.63 & 9.91 & 9.08 \\
SDeMorph (2023) \cite{ref18} & 8.99 & 9.54 & 9.60 & 9.45 & 8.97 & 8.74 \\
Face Demorphing (2022) \cite{ref51} & 9.68 & 10.35 & 10.44 & 10.20 & 10.13 & 9.51 \\
\bottomrule
\end{tabular}
}
\end{table}

We evaluate the proposed MLLM-guided coupled diffusion framework from two complementary perspectives:
(i) \emph{identity restoration performance} using Restoration Accuracy (RA) under multiple FMR operating points, and 
(ii) \emph{image reconstruction fidelity} using PSNR and SSIM metrics. 
All comparisons include landmark-based morphs (AMSL, OpenCV, FaceMorph, WebMorph), diffusion-based morphs (MorDiff), and generative latent-space morphs (StyleGAN).

\textbf{\textit{Identity Restoration Performance:}}
Table~\ref{tab:restoration_accuracy} reports Restoration Accuracy (RA) using ArcFace and AdaFace at 10\%, 1\%, and 0.1\% FMR, where lower FMR corresponds to stricter biometric operating conditions representative of practical deployments. At 0.1\% FMR, the proposed framework maintains robust identity restoration on landmark-based morphs, with the Qwen3-VL variant surpassing 96\% RA on AMSL, OpenCV, and FMorph under both matchers, and reaching 96.35\% on OpenCV with ArcFace. StyleGAN morphs remain the most challenging due to latent-space identity entanglement; nevertheless, under ArcFace at 0.1\% FMR, the Qwen3-VL variant achieves 9.13\% RA, with a similar trend observed for AdaFace (5.34\%), representing a substantial improvement over prior GAN and diffusion-based reference-free demorphing methods such as lc-GAN (2.12\% / 4.52\%), diffDeMorph, IPD, and SDeMorph. At 1\% FMR, performance on StyleGAN further increases to 32.77\% (ArcFace) and 23.86\% (AdaFace), underscoring the effectiveness of semantic MLLM conditioning for disentangling identities embedded in generative latent manifolds of the diffusion model. Finally, at 10\% FMR, our method achieves near-saturation performance across all landmark-based datasets and MorDiff, with the Qwen3-VL guided variant consistently exceeding 99.7\% RA under both face matchers on all non-StyleGAN morphs, indicating strong preservation of identity-discriminative features. Qualitative results are shown in Figure~\ref{fig:main-result} and Figure~\ref{fig:main-result2}.
\begin{table}[t]
\centering
\caption{Demorphing evaluated as top-K retrieval. 
Mean performance across six datasets. All values are reported in percentage (\%).}
\label{tab:retrieval_adaface}
\begin{tabular}{lccc}
\toprule
\textbf{Method} & \textbf{mAP@1} & \textbf{mAP@10} & \textbf{R@10} \\
\midrule
\textbf{\methodname} & \textbf{94.67} & \textbf{93.98} & \textbf{97.65} \\
DiffDeMorph         & 92.59          & 92.76          & 96.43          \\
LCGAN               & 93.24          & 92.78          & 96.53          \\
DCGAN               & 91.68          & 92.59          & 95.89          \\
IPD                 & 89.45          & 86.66          & 94.13          \\
SDeMorph            & 84.60          & 82.69          & 89.45          \\
\bottomrule
\end{tabular}
\end{table}

\textbf{\textit{Retrieval Performance:}}
In addition to standard verification metrics, we evaluate demorphing as a retrieval task by performing nearest-neighbor search in the matcher embedding space with $k \in \{1,10\}$, and report Recall@$k$ and mAP@$k$ following standard retrieval definitions. Positive recall at rank $k$ is defined as the presence of at least one true source identity among the top-$k$ retrieved results. Mean Average Precision at rank $k$ (mAP@$k$) is computed by averaging the precision at each relevant rank up to $k$, and then averaging over all queries. In the context of demorphing, mAP@1 measures strict top-rank identity recovery, while mAP@10 reflects ranking quality within the top-10 candidates and Recall@10 measures whether the correct identities are retrieved within the top-10 results. To increase retrieval difficulty and better simulate a realistic open-set scenario, we augment the gallery with 10,000 face images from the CASIA-WebFace dataset~\cite{yi2014learning}, resulting in a total gallery size of 10,102 entries (10,000 distractors $+$ 102 subject images).

We report the results in Table~\ref{tab:retrieval_adaface}. \methodname\ achieves the strongest overall performance, with 94.67\% mAP@1, 93.98\% mAP@10, and 97.65\% R@10, outperforming all competing methods across datasets. The most pronounced improvement occurs on the challenging StyleGAN morphs, where \methodname\ attains 68.49\% mAP@1, surpassing LCGAN by $+$8.18\% and DiffDeMorph by $+$12.35\%. This substantial gain on GAN-generated morphs indicates improved robustness to synthetic blending artifacts and stronger recovery of identity-discriminative cues. The consistent improvements across both precision-oriented (mAP) and recall-oriented metrics demonstrate that our method not only retrieves the correct identities at top rank but also maintains a coherent identity neighborhood structure.

\textbf{\textit{Image Reconstruction Fidelity:}}
\noindent Beyond identity restoration, we evaluate pixel-level reconstruction quality using PSNR and SSIM (Table~\ref{tab:psnr}). Our method substantially outperforms prior approaches in PSNR, with Qwen3-VL achieving the highest scores on five of six benchmarks, reaching 19.56~dB on StyleGAN and 18.55~dB on MorDiff and exceeding lc-GAN by over +6~dB on average and SDeMorph by nearly +9~dB. Detailed results are presented in Table \ref{tab:iqa_llava} and \ref{tab:iqa_qwen}. However, both metrics operate on pixel-level statistics without capturing semantic or identity-level information; a reconstruction may score poorly while still correctly recovering the target identity. Results should therefore be interpreted alongside the identity-aware metrics reported above.

\textbf{\textit{Effect of MLLM Conditioning:} }Comparing Qwen3-VL and LLaVA-1.6 reveals consistent benefits from stronger multimodal reasoning. While both variants perform similarly on landmark-based morphs at relaxed thresholds, the Qwen-conditioned model demonstrates clear advantages under stricter FMR settings and on generative morphs.
For example, on StyleGAN at 1\% FMR (ArcFace), Qwen3-VL achieves 32.77\% RA compared to 23.97\% with LLaVA-1.6. PSNR improvements are also consistently higher with Qwen conditioning. These results indicate that richer semantic embeddings provide more effective guidance for disentangling identity information during diffusion-based reconstruction.

\section{Ablation Study}
\label{sec:ablation}

\begin{table*}[htbp]
\centering
\caption{Ablation study of Qwen3-VL components for demorphing. Restoration Accuracy measures the contribution of each component. Unless specified as ``Latent'', all experiments are conducted in RGB space.}
\label{tab:ablation_qwen}
\resizebox{\textwidth}{!}{%
\begin{tabular}{@{}c|c|cccccc|cccccc@{}}
\toprule
\multirow{2}{*}{\textbf{Component}} & \multirow{2}{*}{\textbf{RA}} & \multicolumn{6}{c|}{\textbf{ArcFace}} & \multicolumn{6}{c}{\textbf{AdaFace}} \\
\cmidrule(lr){3-8} \cmidrule(lr){9-14}
& & AMSL & OpenCV & FMorph & MorDiff & Wmorph & StyleGAN & AMSL & OpenCV & FMorph & MorDiff & Wmorph & StyleGAN \\
\midrule
\multirow{3}{*}{Initial Layer} 
& 10\% & 99.88\% & 100.00\% & 99.60\% & 100.00\% & 99.87\% & 67.25\% & 99.83\% & 100.00\% & 100.00\% & 100.00\% & 99.96\% & 59.32\% \\ 
& 1\%  & 99.31\% & 99.87\% & 98.79\% & 100.00\% & 98.82\% & 34.71\% & 99.39\% & 99.69\% & 99.46\% & 100.00\% & 99.34\% & 28.22\% \\ 
& 0.1\% & 96.64\% & 98.62\% & 96.77\% & 91.54\% & 95.20\% & 11.06\% & 96.72\% & 98.94\% & 97.85\% & 93.49\% & 95.85\% & 7.27\% \\
\midrule
\multirow{3}{*}{Middle Layer} 
& 10\% & 99.98\% & 100.00\% & 99.93\% & 98.53\% & 99.78\% & 67.23\% & 99.74\% & 100.00\% & 100.00\% & 99.79\% & 99.91\% & 57.28\% \\ 
& 1\%  & 99.35\% & 99.87\% & 99.14\% & 98.53\% & 99.14\% & 32.77\% & 98.95\% & 99.96\% & 99.73\% & 98.42\% & 99.27\% & 23.86\% \\ 
& 0.1\% & 96.35\% & 98.65\% & 96.55\% & 94.63\% & 95.22\% & 9.13\% & 96.41\% & 98.74\% & 98.14\% & 95.37\% & 96.25\% & 5.34\% \\
\midrule
\multirow{3}{*}{Last Layer} 
& 10\% & 99.98\% & 99.91\% & 100.00\% & 99.57\% & 99.83\% & 65.35\% & 99.95\% & 99.96\% & 100.00\% & 99.89\% & 99.96\% & 55.94\% \\
& 1\%  & 99.43\% & 99.70\% & 99.54\% & 97.76\% & 98.97\% & 30.96\% & 99.69\% & 99.96\% & 99.61\% & 98.61\% & 99.57\% & 19.95\% \\
& 0.1\% & 96.38\% & 98.66\% & 97.04\% & 91.88\% & 95.03\% & 8.80\% & 97.56\% & 99.14\% & 98.55\% & 94.12\% & 96.57\% & 4.34\% \\
\midrule
\midrule
\multirow{3}{*}{\makecell{Latent + \\Initial Layer}} 
& 10\% & 96.82\% & 97.96\% & 97.30\% & 95.60\% & 97.27\% & 41.67\% & 92.80\% & 95.13\% & 94.54\% & 93.35\% & 94.83\% & 26.14\% \\ 
& 1\%  & 86.02\% & 89.09\% & 87.12\% & 85.58\% & 87.22\% & 13.30\% & 75.24\% & 80.77\% & 76.38\% & 78.43\% & 77.67\% & 5.84\% \\ 
& 0.1\% & 59.55\% & 66.36\% & 59.36\% & 60.94\% & 60.63\% & 2.78\% & 42.66\% & 51.58\% & 44.22\% & 50.00\% & 46.61\% & 1.04\% \\
\midrule
\multirow{3}{*}{\makecell{Latent + \\Middle Layer}} 
& 10\% & 96.17\% & 97.22\% & 96.68\% & 94.82\% & 96.04\% & 38.08\% & 90.57\% & 93.77\% & 91.34\% & 92.68\% & 91.34\% & 23.01\% \\ 
& 1\%  & 80.90\% & 86.34\% & 82.62\% & 81.91\% & 81.97\% & 10.76\% & 67.55\% & 74.00\% & 67.50\% & 74.29\% & 68.77\% & 4.76\% \\ 
& 0.1\% & 50.02\% & 56.94\% & 52.38\% & 58.84\% & 51.90\% & 1.90\% & 34.29\% & 41.53\% & 34.57\% & 47.36\% & 36.84\% & 0.87\% \\
\midrule
\multirow{3}{*}{\makecell{Latent + \\Last Layer}} 
& 10\% & 96.02\% & 97.62\% & 96.23\% & 97.76\% & 96.40\% & 35.89\% & 91.86\% & 94.70\% & 93.53\% & 98.93\% & 92.52\% & 22.72\% \\ 
& 1\%  & 82.40\% & 86.95\% & 83.17\% & 83.67\% & 83.22\% & 9.35\% & 69.40\% & 78.86\% & 71.17\% & 75.61\% & 74.09\% & 4.64\% \\ 
& 0.1\% & 51.43\% & 59.09\% & 52.07\% & 58.16\% & 54.38\% & 1.86\% & 36.82\% & 45.87\% & 37.69\% & 46.33\% & 39.46\% & 0.83\% \\
\midrule
\midrule
\multirow{3}{*}{ViT features} 
& 10\% & 95.52\% & 98.63\% & 98.05\% & 97.45\% & 95.57\% & 50.09\% & 93.33\% & 98.27\% & 97.57\% & 97.45\% & 94.11\% & 36.55\% \\ 
& 1\%  & 82.66\% & 92.07\% & 88.68\% & 90.11\% & 82.29\% & 17.54\% & 78.36\% & 89.86\% & 86.86\% & 91.17\% & 79.63\% & 9.97\% \\ 
& 0.1\% & 57.45\% & 71.52\% & 67.12\% & 76.70\% & 57.09\% & 3.78\% & 52.07\% & 69.44\% & 64.89\% & 76.28\% & 53.41\% & 1.69\% \\
\midrule
\multirow{3}{*}{\makecell{Latent + \\ViT features}} 
& 10\% & 97.02\% & 97.92\% & 98.11\% & 95.93\% & 97.16\% & 41.45\% & 91.90\% & 94.67\% & 93.89\% & 94.09\% & 93.82\% & 24.32\% \\ 
& 1\%  & 84.53\% & 88.88\% & 87.03\% & 86.86\% & 85.00\% & 11.78\% & 70.59\% & 76.88\% & 73.87\% & 78.82\% & 73.78\% & 4.85\% \\ 
& 0.1\% & 54.36\% & 63.25\% & 59.13\% & 61.00\% & 57.13\% & 2.03\% & 37.70\% & 45.33\% & 42.88\% & 48.47\% & 39.82\% & 0.75\% \\
\bottomrule
\end{tabular}%
}
\end{table*}
\begin{table*}[htbp]
\centering
\caption{Ablation study of LLaVA-1.6 components for demorphing. Restoration Accuracy measures the contribution of each component. Unless specified as ``Latent'', all experiments are conducted in RGB space.}
\label{tab:ablation_llava}
\resizebox{\textwidth}{!}{%
\begin{tabular}{@{}c|c|cccccc|cccccc@{}}
\toprule
\multirow{2}{*}{\textbf{Component}} & \multirow{2}{*}{\textbf{RA}} & \multicolumn{6}{c|}{\textbf{ArcFace}} & \multicolumn{6}{c}{\textbf{AdaFace}} \\
\cmidrule(lr){3-8} \cmidrule(lr){9-14}
& & AMSL & OpenCV & FMorph & MorDiff & Wmorph & StyleGAN & AMSL & OpenCV & FMorph & MorDiff & Wmorph & StyleGAN \\
\midrule
\multirow{3}{*}{Initial Layer} 
& 10\% & 99.03\% & 99.64\% & 99.43\% & 99.07\% & 98.94\% & 58.37\% & 98.92\% & 99.53\% & 99.51\% & 99.46\% & 98.97\% & 47.83\% \\ 
& 1\%  & 95.43\% & 97.34\% & 96.67\% & 95.82\% & 95.38\% & 23.74\% & 94.96\% & 97.63\% & 97.54\% & 96.57\% & 94.72\% & 15.36\% \\ 
& 0.1\% & 84.83\% & 90.07\% & 86.54\% & 85.96\% & 83.58\% & 6.24\% & 83.94\% & 91.53\% & 89.63\% & 88.14\% & 82.97\% & 2.94\% \\
\midrule
\multirow{3}{*}{Middle Layer} 
& 10\% & 99.83\% & 99.84\% & 99.87\% & 99.76\% & 99.72\% & 56.24\% & 99.73\% & 99.91\% & 99.82\% & 99.67\% & 99.53\% & 46.74\% \\ 
& 1\%  & 97.14\% & 99.23\% & 98.56\% & 97.52\% & 97.48\% & 23.97\% & 97.63\% & 99.24\% & 99.31\% & 97.68\% & 97.43\% & 17.04\% \\ 
& 0.1\% & 88.82\% & 94.63\% & 92.54\% & 87.71\% & 87.74\% & 6.62\% & 90.91\% & 96.82\% & 94.37\% & 90.73\% & 90.53\% & 2.94\% \\
\midrule
\multirow{3}{*}{Last Layer} 
& 10\% 
& 99.47\% & 99.93\% 
& 99.73\% & 99.54\% & 99.43\% & 57.46\% 
& 99.63\% & 100.00\% 
& 99.72\% & 99.63\% & 99.34\% & 47.43\% \\
& 1\%  
& 97.34\% & 99.24\% 
& 97.73\% & 96.83\% & 96.54\% & 24.07\% 
& 97.43\% & 99.73\% 
& 98.53\% & 97.24\% & 96.23\% & 16.37\% \\
& 0.1\% 
& 88.73\% & 94.53\% 
& 89.63\% & 86.94\% & 85.73\% & 6.53\% 
& 91.07\% & 97.13\% 
& 92.14\% & 89.53\% & 86.83\% & 3.04\% \\
\midrule

\midrule
\multirow{3}{*}{\makecell{Latent + \\Initial Layer}} 
& 10\% & 77.83\% & 81.74\% & 79.13\% & 81.53\% & 80.73\% & 23.07\% & 66.73\% & 71.53\% & 69.13\% & 73.73\% & 71.53\% & 12.73\% \\ 
& 1\%  & 46.34\% & 49.07\% & 45.23\% & 55.43\% & 48.83\% & 5.63\% & 32.53\% & 38.53\% & 31.63\% & 45.07\% & 37.03\% & 2.63\% \\ 
& 0.1\% & 15.63\% & 17.43\% & 15.43\% & 22.23\% & 18.34\% & 0.73\% & 8.34\% & 10.93\% & 8.13\% & 15.13\% & 10.73\% & 0.23\% \\
\midrule
\multirow{3}{*}{\makecell{Latent + \\Middle Layer}} 
& 10\% & 93.23\% & 95.03\% & 92.83\% & 92.53\% & 94.34\% & 35.93\% & 86.43\% & 90.43\% & 87.83\% & 88.43\% & 88.93\% & 19.83\% \\ 
& 1\%  & 74.73\% & 78.93\% & 73.73\% & 76.63\% & 75.34\% & 10.53\% & 60.63\% & 66.93\% & 60.83\% & 68.13\% & 63.34\% & 4.23\% \\ 
& 0.1\% & 42.34\% & 47.93\% & 42.13\% & 48.03\% & 44.53\% & 1.93\% & 27.83\% & 32.43\% & 26.43\% & 38.23\% & 29.93\% & 0.63\% \\
\midrule
\multirow{3}{*}{\makecell{Latent + \\Last Layer}} 
& 10\% & 72.43\% & 74.53\% & 72.63\% & 76.73\% & 71.23\% & 19.43\% & 56.23\% & 61.63\% & 58.73\% & 69.03\% & 58.23\% & 8.53\% \\ 
& 1\%  & 36.93\% & 39.63\% & 37.34\% & 49.23\% & 38.73\% & 4.03\% & 22.73\% & 26.13\% & 21.93\% & 35.73\% & 24.83\% & 0.93\% \\ 
& 0.1\% & 11.23\% & 13.13\% & 12.23\% & 20.73\% & 13.73\% & 0.63\% & 4.83\% & 6.83\% & 5.33\% & 13.23\% & 6.43\% & 0.23\% \\
\midrule

\midrule
\multirow{3}{*}{ViT features} 
& 10\% & 99.13\% & 99.63\% & 99.43\% & 99.34\% & 99.13\% & 56.43\% & 98.53\% & 99.63\% & 99.34\% & 99.34\% & 98.73\% & 44.63\% \\ 
& 1\%  & 94.73\% & 98.03\% & 96.53\% & 96.13\% & 93.43\% & 22.93\% & 93.63\% & 97.34\% & 96.83\% & 95.63\% & 93.83\% & 13.13\% \\ 
& 0.1\% & 80.53\% & 89.83\% & 88.13\% & 84.83\% & 79.83\% & 6.23\% & 79.03\% & 89.23\% & 88.73\% & 85.53\% & 80.63\% & 2.63\% \\
\midrule
\multirow{3}{*}{\makecell{Latent + \\ViT features}} 
& 10\% & 97.23\% & 97.93\% & 97.03\% & 96.63\% & 96.93\% & 40.73\% & 93.53\% & 95.63\% & 92.73\% & 92.93\% & 94.73\% & 26.13\% \\ 
& 1\%  & 84.93\% & 88.13\% & 83.93\% & 85.03\% & 86.23\% & 12.63\% & 74.93\% & 79.63\% & 73.93\% & 81.23\% & 76.63\% & 6.03\% \\ 
& 0.1\% & 57.03\% & 60.93\% & 55.63\% & 64.83\% & 58.93\% & 2.33\% & 44.13\% & 47.23\% & 43.83\% & 53.43\% & 46.03\% & 0.73\% \\
\bottomrule
\end{tabular}%
}
\end{table*}

\begin{table*}[ht]
\centering
\caption{Ablation study on Qwen3-VL components, reporting reconstruction performance using PSNR and SSIM across the tested datasets.}
\label{tab:iqa_qwen}
\begin{tabular}{|l|c|c|c|c|c|c|}
\hline
\textbf{Component} & \textbf{AMSL} & \textbf{FMorph} & \textbf{OpenCV} & \textbf{StyleGAN} & \textbf{WMorph} &\textbf{MorDiff}\\
\hline
Initial Layer 
& 17.32 / 0.40 
& 17.07 / 0.41 
& 17.15 / 0.42 
& 17.66 / 0.48 
& 17.11 / 0.38 
& 17.22 / 0.42 \\
\hline
Middle Layer 
& 18.36 / 0.39 
& 18.26 / 0.40 
& 18.34 / 0.41 
& 19.56 / 0.46 
& 17.99 / 0.36 
& 18.55 / 0.41 \\
\hline
Last Layer 
& 17.72 / 0.40 
& 17.44 / 0.41 
& 17.58 / 0.42 
& 18.37 / 0.49 
& 17.30 / 0.38 
& 17.88 / 0.41 \\
\hline
\hline
Latent + Initial Layer
& 16.13 / 0.51 
& 15.66 / 0.47 
& 15.95 / 0.47 
& 16.56 / 0.49 
& 15.75 / 0.49 
& 16.31 / 0.48 \\
\hline
Latent + Last Layer
& 16.16 / 0.49 
& 15.33 / 0.45 
& 15.64 / 0.45 
& 16.79 / 0.47 
& 15.67 / 0.48 
& 15.82 / 0.48 \\
\hline
Latent + Middle Layer 
& 16.69 / 0.51 
& 15.88 / 0.46 
& 16.19 / 0.46 
& 17.40 / 0.48 
& 16.09 / 0.49 
& 16.43 / 0.48 \\
\hline
\hline
ViT features 
& 16.02 / 0.38 
& 15.95 / 0.39 
& 15.90 / 0.39 
& 16.09 / 0.42 
& 16.02 / 0.37 
& 15.90 / 0.38 \\
\hline
Latent + ViT features 
& 16.00 / 0.49 
& 15.48 / 0.45 
& 15.70 / 0.44 
& 16.39 / 0.46 
& 15.58 / 0.47 
& 15.88 / 0.47 \\
\hline
\end{tabular}
\end{table*}

\begin{table*}[ht]
\centering
\caption{Ablation study on LLaVA-1.6 components, reporting reconstruction performance using PSNR and SSIM across the tested datasets.}
\label{tab:iqa_llava}

\begin{tabular}{|l|c|c|c|c|c|c|}
\hline
\textbf{Component} & \textbf{AMSL} & \textbf{FMorph} & \textbf{OpenCV} & \textbf{StyleGAN} & \textbf{WMorph} & \textbf{MorDiff} \\
\hline
Initial Layer 
& 17.32 / 0.41 
& 17.34 / 0.41 
& 17.22 / 0.42 
& 18.01 / 0.47 
& 17.14 / 0.39 
& 17.35 / 0.43 \\
\hline
Middle Layer 
& 16.13 / 0.38 
& 16.17 / 0.39 
& 16.32 / 0.39 
& 16.06 / 0.44 
& 16.11 / 0.36 
& 16.18 / 0.42 \\
\hline
Last Layer 
& 16.80 / 0.40 
& 16.73 / 0.40 
& 16.66 / 0.42 
& 16.73 / 0.40 
& 16.72 / 0.39 
& 16.73 / 0.40 \\
\hline
\hline
Latent + Initial Layer 
& 16.07 / 0.51 
& 15.44 / 0.46 
& 15.71 / 0.46 
& 16.48 / 0.47 
& 15.69 / 0.50 
& 15.54 / 0.53 \\
\hline
Latent + Last Layer 
& 16.08 / 0.48 
& 15.48 / 0.44 
& 15.74 / 0.43 
& 16.40 / 0.42 
& 15.58 / 0.47 
& 15.98 / 0.53 \\
\hline
Latent + Middle Layer 
& 16.12 / 0.49 
& 15.35 / 0.45 
& 15.67 / 0.44 
& 16.25 / 0.46 
& 15.70 / 0.48 
& 15.95 / 0.53 \\
\hline
\hline
ViT features 
& 17.01 / 0.39 
& 16.86 / 0.39 
& 16.85 / 0.40 
& 17.35 / 0.44 
& 16.98 / 0.38 
& 17.17 / 0.42 \\
\hline
Latent + ViT features 
& 16.06 / 0.49 
& 15.42 / 0.44 
& 15.59 / 0.44 
& 16.63 / 0.46 
& 15.55 / 0.47 
& 15.99 / 0.54 \\
\hline
\end{tabular}
\end{table*}

We perform a detailed ablation analysis to examine: 
(1) the effect of MLLM depth, 
(2) demorphing in RGB versus latent space, 
(3) the impact of using ViT features instead of full MLLM embeddings,  
(4)  the impact of using MLLM hidden states as the conditional signal and (5) the effect of prompting strategy.
Results are summarized in Tables~\ref{tab:ablation_qwen}, \ref{tab:ablation_llava}, \ref{tab:iqa_qwen},\ref{tab:iqa_llava} and \ref{tab:ablation_text}.

\textbf{\textit{Effect of MLLM Depth:}}
We analyze which internal layer of the MLLM provides the most identity-discriminative representation for demorphing. Embeddings are extracted from the \textit{Initial}, \textit{Middle}, and \textit{Last} transformer blocks. Under the strictest restoration accuracy threshold (RA = 0.1\%), no single layer consistently dominates across both models. For Qwen-VL, performance differences between layers are small and inconsistent across datasets, with no clear ordering between Initial, Middle, and Last layers. For LLaVA-1.6, the Middle layer outperforms both Initial and Last layers across most datasets and evaluation metrics. Overall, \textbf{the middle layer represents the most suitable and reliable choice} across both models and face recognition systems, suggesting that intermediate transformer representations encode sufficiently structured identity features for precise demorphing without overfitting to task-specific patterns in deeper layers. Across both models, performance on StyleGAN-based morphs remains substantially lower than on classical morph types regardless of layer depth, indicating that GAN-generated morphs are inherently more challenging to reverse. Detailed results are presented in Table~\ref{tab:ablation_qwen} and Table~\ref{tab:ablation_llava}.

\textbf{\textit{RGB vs.\ Latent Domain Demorphing:}}
We next compare demorphing performed directly in the RGB feature space with variants that incorporate latent-space representations.

Across both models, RGB-based demorphing significantly outperforms latent-conditioned variants. 
When operating purely in RGB space, restoration accuracy remains near-saturated at FMR = 10\% and stays high even at FMR = 1\%. 
In contrast, incorporating latent representations (``Latent + Layer'' variants) results in a substantial degradation in performance.

For Qwen3-VL, performance drops markedly at FMR = 0.1\%, with reductions of up to 30--40\% compared to RGB-only embeddings. 
The degradation is even more pronounced for LLaVA-1.6, where decreases exceeding 40\% are observed under strict operating points.

These findings indicate that demorphing is fundamentally an image reconstruction task that relies on fine-grained spatial and texture cues. 
Latent representations tend to entangle identity information  and suppress high-frequency facial details, thereby reducing identity and image separability. Moreover, reconstruction back to the RGB domain from latent representations is inherently constrained by the capacity and inductive biases of the decoder, which may fail to faithfully recover fine-grained identity cues lost during encoding. Overall, \textbf{demorphing in the RGB domain is significantly more reliable than in latent space}.

\begin{table*}[htbp]
\centering
\caption{Effect of text encoding and re-encoding on demorphing performance. Re-encoding textual representations leads to significant restoration accuracy degradation. {\textcolor{dropred}{$\downarrow$}} indicates absolute drop in percentage points from the original method.}
\label{tab:ablation_text}
\resizebox{\textwidth}{!}{%
\begin{tabular}{@{}c|c|cccccc|cccccc@{}}
\toprule
\multirow{2}{*}{\textbf{Component}} & \multirow{2}{*}{\textbf{RA}} & \multicolumn{6}{c|}{\textbf{ArcFace}} & \multicolumn{6}{c}{\textbf{AdaFace}} \\
\cmidrule(lr){3-8} \cmidrule(lr){9-14}
& & AMSL & OpenCV & FMorph & MorDiff & Wmorph & StyleGAN & AMSL & OpenCV & FMorph & MorDiff & Wmorph & StyleGAN \\
\midrule
\multirow{3}{*}{Qwen3-VL}
& @ 10\% FMR
& \cellval{89.40}{10.48} & \cellval{95.39}{4.61} & \cellval{96.10}{3.50} & \cellval{93.24}{6.76} & \cellval{92.31}{7.56} & \cellval{38.26}{28.99}
& \cellval{85.95}{13.88} & \cellval{93.41}{6.59} & \cellval{93.98}{6.02} & \cellval{91.13}{8.83} & \cellval{89.29}{10.67} & \cellval{27.39}{31.93} \\
& @ 1\% FMR
& \cellval{69.17}{30.14} & \cellval{81.05}{18.82} & \cellval{81.05}{17.74} & \cellval{77.49}{22.51} & \cellval{73.51}{25.31} & \cellval{11.74}{22.97}
& \cellval{64.10}{35.29} & \cellval{78.18}{21.51} & \cellval{78.18}{21.28} & \cellval{72.84}{27.16} & \cellval{66.35}{32.99} & \cellval{6.91}{21.31} \\
& @ 0.1\% FMR
& \cellval{40.81}{55.83} & \cellval{55.42}{43.20} & \cellval{53.76}{43.01} & \cellval{52.44}{39.10} & \cellval{43.88}{51.32} & \cellval{2.61}{8.45}
& \cellval{33.07}{63.65} & \cellval{51.21}{47.73} & \cellval{49.18}{48.67} & \cellval{48.56}{44.93} & \cellval{37.11}{58.74} & \cellval{1.13}{6.14} \\
\midrule
\multirow{3}{*}{LLaVA-1.6}
& @ 10\% FMR
& \cellval{90.69}{9.12} & \cellval{93.94}{5.89} & \cellval{92.98}{6.83} & \cellval{90.65}{9.08} & \cellval{90.64}{9.09} & \cellval{34.82}{21.41}
& \cellval{82.15}{17.56} & \cellval{88.03}{11.94} & \cellval{84.37}{14.91} & \cellval{85.75}{13.80} & \cellval{81.00}{18.55} & \cellval{21.36}{25.41} \\
& @ 1\% FMR
& \cellval{69.68}{27.44} & \cellval{72.54}{26.70} & \cellval{72.87}{25.68} & \cellval{72.94}{24.46} & \cellval{68.64}{28.76} & \cellval{10.06}{13.91}
& \cellval{55.09}{42.55} & \cellval{61.31}{37.96} & \cellval{58.26}{40.87} & \cellval{62.81}{34.83} & \cellval{54.08}{43.56} & \cellval{4.85}{12.15} \\
& @ 0.1\% FMR
& \cellval{35.55}{53.31} & \cellval{40.89}{53.75} & \cellval{40.84}{51.74} & \cellval{43.54}{44.20} & \cellval{34.90}{52.81} & \cellval{1.68}{4.99}
& \cellval{24.82}{66.09} & \cellval{30.89}{65.95} & \cellval{26.31}{68.42} & \cellval{33.96}{56.29} & \cellval{24.51}{65.74} & \cellval{0.53}{2.37} \\
\bottomrule
\end{tabular}%
}
\end{table*}

\textbf{\textit{ViT Features vs.\ MLLM Embeddings:}}
We compare full MLLM embeddings with features extracted solely from the underlying ViT backbone. Although ViT features remain competitive at relaxed operating points (FMR = 10\%), their performance degrades more rapidly under stricter thresholds. 
At FMR = 0.1\%, MLLM embeddings consistently outperform ViT features across both ArcFace and AdaFace. 
This demonstrates that \textbf{multimodal transformer layers provide additional identity-discriminative abstraction beyond raw visual features required to disentangle hard GAN based morphs}.

The advantage of MLLM embeddings likely stems from higher-level semantic structuring and improved feature disentanglement introduced through multimodal pretraining. 
Nevertheless, the strong performance of ViT-only features indicates that high-quality visual representations already encode substantial identity information which are further enriched by the MLLM transformer layers.

\textbf{\textit{Hidden States vs.\ Text Conditioning:}}
To analyze the effect of the conditioning signal, we replace multimodal hidden-state conditioning with textual descriptions generated by the MLLM. The generated text is encoded using CLIP~\cite{radford2021learning}, and the resulting embedding is used as the conditioning input to the diffusion model.
As shown in Table~\ref{tab:ablation_text}, text-based conditioning consistently reduces restoration accuracy across all datasets. While the drop at $10\%$ FMR is moderate (e.g., $96.10\% \rightarrow 92.98\%$), performance degrades substantially at stricter operating points, particularly at $1\%$ and $0.1\%$ FMR (e.g., $78.18\% \rightarrow 58.26\%$). In contrast to the near-saturated results in Table~\ref{tab:restoration_accuracy}, the re-encoded variant collapses under low-FMR constraints.  These findings indicate that \textbf{hidden states provide a richer, identity-preserving conditioning signal}, whereas textual conditioning introduces an information bottleneck that suppresses fine-grained identity cues critical for robust demorphing.

\textbf{\textit{Prompt Sensitivity Analysis:}}
To evaluate the sensitivity of our method to prompt design, we compare demorphing performance under a task-specific prompt versus a generic prompt. The task-specific prompt instructs the model with a forensic framing: \texttt{``You are a forensic facial morphing analyst. Your job is to detect and analyze face images that may be composites or morphs of two different individuals''}, with the user query \texttt{``Analyze this face image for signs of morphing, blending artifacts, or dual identity signals''}. In contrast, the generic prompt simply defines the model as \texttt{``You are a forensic expert and your job is to identify faces.''}, with the user query \texttt{``Describe the image.''}. When averaged across all morph attacks at 0.1\% FMR, the generic prompt achieves a mean Restoration Accuracy of 79.66\% and 79.33\% under ArcFace and AdaFace respectively, outperforming the task-specific prompt which yields 77.97\% and 77.86\%, an improvement of +1.69\% and +1.47\% respectively. The largest absolute gap is observed on the StyleGAN attack, where the generic prompt yields 11.06\% versus 8.44\% for the task-specific prompt at 0.1\% FMR, suggesting that GAN-based morphs are the most sensitive to prompt phrasing. Overall, this behaviour suggests that overly specific prompts may inadvertently constrain the model's feature extraction by biasing attention toward task-relevant regions, whereas a generic prompt allows the model to leverage its broader visual understanding more freely.

\begin{table}[t]
\centering
\caption{Comparison of morph consistency and inter-identity separation across datasets using AdaFace and ArcFace embeddings. M-BF1 and M-BF2 denote morph-to-identity mean similarity scores, while BF1-BF2 captures separation between the two constituent identities.}

\resizebox{\linewidth}{!}{%
\begin{tabular}{lcccccc}
\toprule
& \multicolumn{3}{c}{AdaFace} & \multicolumn{3}{c}{ArcFace} \\
\cmidrule(lr){2-4} \cmidrule(lr){5-7}
Dataset & M-BF1 & M-BF2 & BF1-BF2 & M-BF1 & M-BF2 & BF1-BF2 \\
\midrule
AMSL      & 0.62 & 0.46 & 0.03 & 0.63 & 0.44 & 0.06 \\
FaceMorph & 0.54 & 0.54 & 0.03 & 0.52 & 0.52 & 0.06 \\
MorDiff   & 0.49 & 0.49 & 0.09 & 0.48 & 0.48 & 0.11 \\
OpenCV    & 0.55 & 0.55 & 0.03 & 0.54 & 0.53 & 0.05 \\
\rowcolor{gray!20}
StyleGAN  & 0.20 & 0.19 & 0.03 & 0.21 & 0.20 & 0.05 \\
WebMorph  & 0.55 & 0.54 & 0.03 & 0.53 & 0.53 & 0.05 \\
\bottomrule
\end{tabular}%
}
\label{tab:morph_distance}
\end{table}

\textbf{\textit{Failure Analysis:}} StyleGAN morphs exhibit a clear degradation in identity retention, which directly explains the lower performance of our demorphing method. The average Morph–BF1 and Morph–BF2 similarities are both significantly reduced and nearly collapsed (AdaFace: 0.195 and 0.189; ArcFace: 0.211 and 0.204), unlike datasets such as AMSL and FaceMorph where these values remain consistently higher ( $\approx$ 0.52–0.62 for Morph–BF1/BF2). This indicates that StyleGAN morphs fail to preserve a balanced embedding proximity to both constituent identities, instead drifting away from the expected mid-point representation in feature space (see Supplementary for detailed analysis). As a result, our method receives weak and uninformative identity conditioning from the morph image, leading to degraded demorphing performance that is fundamentally attributable to the poor identity fidelity of the morph generation process rather than limitations of the reconstruction model. We present these results in Table \ref{tab:morph_distance}.

\section{Conclusion}

We presented a reference-free facial demorphing approach that leverages MLLMs to guide a coupled diffusion-based reconstruction process. By extracting semantic embeddings from intermediate MLLM layers to condition a denoising diffusion network, our method combines high-level semantic reasoning with fine-grained perceptual information to disentangle morphed images into their constituent images.

Experiments demonstrate state-of-the-art performance across multiple benchmarks. On landmark-based morphs, our method exceeds 96\% restoration accuracy at the stringent 0.1\% FMR operating point using ArcFace and AdaFace. It also generalizes strongly to generative attacks, outperforming prior methods on StyleGAN and diffusion-based morphs in both identity accuracy and perceptual quality (PSNR, SSIM). Ablations show that (1) intermediate MLLM layers encode the most identity-discriminative semantics, (2) RGB-space diffusion outperforms latent-space reconstruction by preserving fine-grained cues, (3) full MLLM embeddings outperform raw ViT features, and (4) direct hidden-state conditioning beats text-based conditioning.

Challenges remain in handling highly entangled generative morphs and reducing the computational cost of MLLM-based conditioning. Future work will explore efficient distillation of semantic embeddings and improved robustness against advanced attacks. Overall, this work highlights the value of integrating foundation model semantics with generative reconstruction for biometric security. Code, embeddings, evaluation scripts, and pretrained models will be released at \texttt{\url{https://huggingface.co/nitishshukla/demorphing_model_weights/}}.



%

\bibliographystyle{ieeetr}
\bibliography{bibliography}

%

\begin{IEEEbiographynophoto}{Nitish Shukla}
 is currently a Ph.D. candidate in the Department of Computer Science and Engineering at Michigan State University. His research interests include facial biometrics and multimodal learning.
\end{IEEEbiographynophoto}
\begin{IEEEbiographynophoto}{Arun Ross}
(Senior Member, IEEE) is the Martin J. Vanderploeg Endowed Professor in Computer Science and Engineering at Michigan State University and Site Director of NSF’s Center for Identification Technology Research (CITeR). Ross is the recipient of the NSF CAREER Award, the IAPR JK Aggarwal Prize, and the IAPR Young Biometrics Investigator Award. His research interests include biometrics, computer vision, and deep learning.
\end{IEEEbiographynophoto}

\ifCLASSOPTIONcaptionsoff
  \newpage
\fi
\end{document}